%% file: PaperForReview.tex
\documentclass[10pt,twocolumn,letterpaper]{article}

\usepackage{cvpr}              

\usepackage{graphicx}
\usepackage{amsmath}
\usepackage{amssymb}
\usepackage{booktabs}
\usepackage{placeins} 
\usepackage{float} 
\usepackage{enumitem}
\usepackage{tabularx} 
\usepackage{longtable} 
\usepackage{array} 
\usepackage{ragged2e} 
\usepackage{booktabs} 
\usepackage{placeins}
\usepackage{afterpage} 
\usepackage{caption} 

\usepackage[pagebackref,breaklinks,colorlinks]{hyperref}

\usepackage[capitalize]{cleveref}
\crefname{section}{Sec.}{Secs.}
\Crefname{section}{Section}{Sections}
\Crefname{table}{Table}{Tables}
\crefname{table}{Tab.}{Tabs.}

\def\cvprPaperID{*****} 
\def\confName{CVPR}
\def\confYear{2022}

\begin{document}

\title{EditID: Training-Free Editable ID Customization for Text-to-Image Generation}


\author{Guandong Li\\
iFlyTek\\
{\tt\small gdli7@iflytek.com}
\and
Zhaobin Chu\\
iFlyTek\\
{\tt\small zbchu2@iflytek.com}
}



\twocolumn[{
\renewcommand\twocolumn[1][]{#1}
\maketitle
\begin{center}
    \centering
    \vspace*{-.8cm}
    \includegraphics[width=\textwidth]{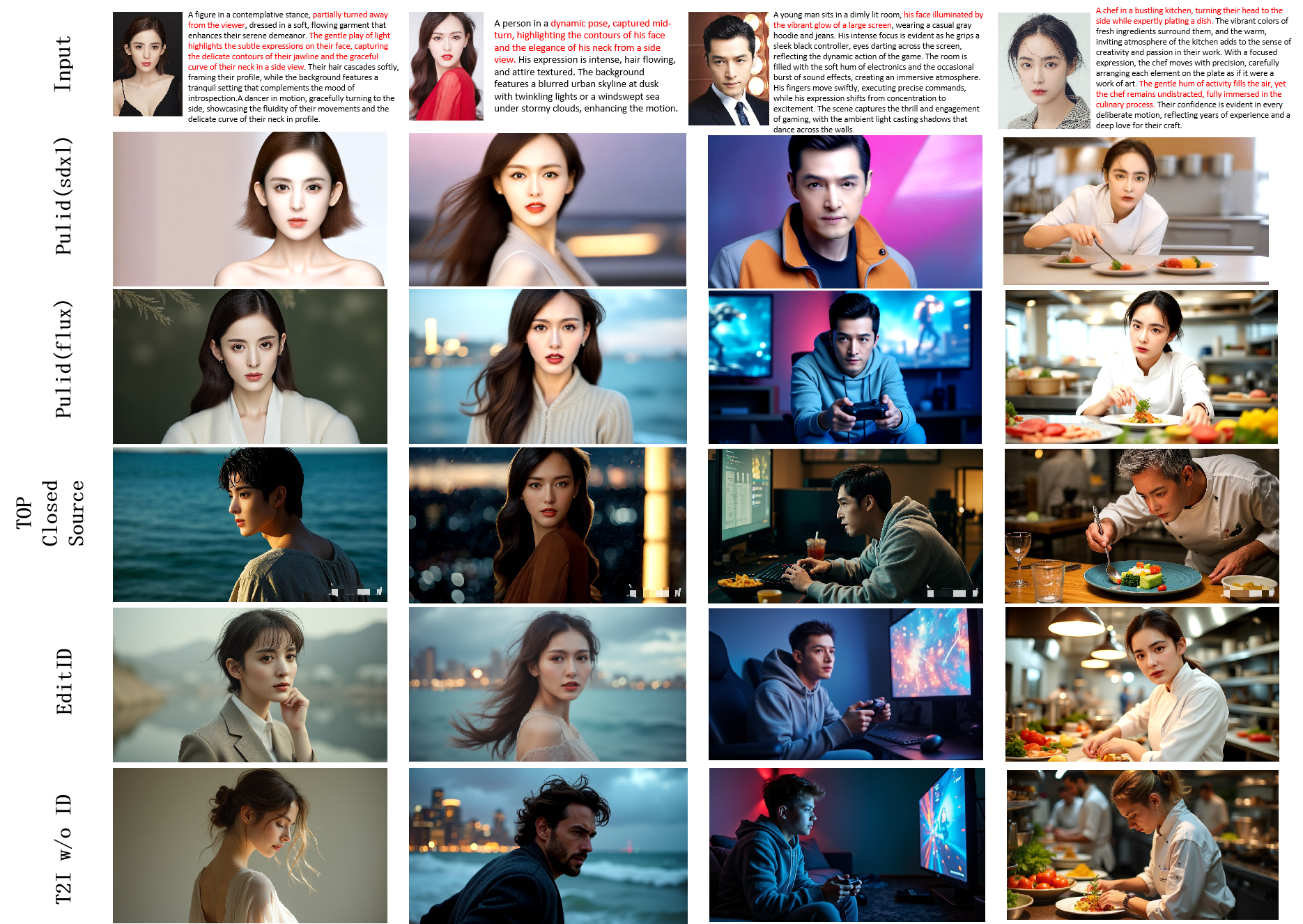}
    \vspace*{-.6cm}
    \captionof{figure}{We introduce EditID, a training-free ID customization approach. EditID achieves better editability compared to similar methods. It demonstrates excellent editability in long prompts (where action prompts are marked in red) and aligns well with Flux's T2I.}
\label{fig:figure1}
\end{center}
}]

\begin{abstract}
We propose EditID, a training-free approach based on the DiT architecture, which achieves highly editable customized IDs for text to image generation. Existing text-to-image models for customized IDs typically focus more on ID consistency while neglecting editability. It is challenging to alter facial orientation, character attributes, and other features through prompts. EditID addresses this by deconstructing the text-to-image model for customized IDs into an image generation branch and a character feature branch. The character feature branch is further decoupled into three modules: feature extraction, feature fusion, and feature integration. By introducing a combination of mapping features and shift features, along with controlling the intensity of ID feature integration, EditID achieves semantic compression of local features across network depths, forming an editable feature space. This enables the successful generation of high-quality images with editable IDs while maintaining ID consistency, achieving excellent results in the IBench evaluation, which is an editability evaluation framework for the field of customized ID text-to-image generation that quantitatively demonstrates the superior performance of EditID. EditID is the first text-to-image solution to propose customizable ID editability on the DiT architecture, meeting the demands of long prompts and high-quality image generation.
\end{abstract}

\section{Introduction}
\label{sec:intro}

ID customization generation \cite{guo2024pulid, gal2022image,kumari2023multi, ruiz2023dreambooth,liu2023facechain}, as a personalized type of Text-to-Image generation, integrates IDs with prompts to create specific appearances. It offers significant application value in scenarios such as story generation and character creation, and it is one of the core selling points of major text-to-image creative production platforms today.

ID customization typically involves three types of methods: fine-tuning \cite{ruiz2023dreambooth, liu2023facechain}, tuning-free \cite{guo2024pulid, xiao2024fastcomposer,ye2023ip,zhang2024flashface}, and train-free \cite{tewel2024training} approaches. Fine-tuning methods require customization for each ID, involving time-consuming and labor-intensive one-to-one training for individual IDs. Tuning-free methods pretrain an ID fusion and integration module on a large portrait dataset, eliminating the need for ID-specific customization during inference. Train-free methods refine existing tuning-free approaches with innovative design, eliminating the need for model retraining during both training and inference phases. Our proposed train-free framework can effectively achieve character ID editability in any ID customization model equipped with a character feature branch.

Current ID customization methods generally prioritize character consistency, thereby overlooking character editability. We define character editability as the ability to generate multidimensional control relative to the input ID in response to changes in text prompts, including variations in facial orientation and limb positioning, as well as the flexible modification of related attributes of the input ID, such as hairstyle, accessories, and even age and gender. Under our definition, current consistency-focused methods generally lack editability. Taking the state-of-the-art consistency generation method PuLID \cite{guo2024pulid} as an example, the model achieves fidelity through ID loss but also introduces semantic and layout losses to control the diversity of the input ID during generation. However, in practical applications, it is nearly impossible to induce significant pose changes in the input ID through prompt words. This prompted us to investigate the reasons behind the loss of editability. We found that the model is essentially performing an ID reconstruction task. During the pre-training of tuning-free methods, the ID and prompt descriptions remain almost identical, and the model leverages feature information from the ID to bias the training parameters toward the distribution of the training set. This implies that the model intends to directly replicate the character’s features. However, during inference, when the prompts and ID are inconsistent, excessively strong feature constraints from the character branch result in the output ID lacking the ability to adapt to changes in the prompt. In some cases, this even leads to a "copy-paste" effect between the input ID and output ID in the facial region, as shown in \cref{fig:figure1}. The core contribution of this paper lies in modulating the control strength of the character branch within a train-free framework, achieving a text-to-image model that significantly enhances ID editability while maintaining ID consistency.

The core of our approach lies in enhancing editability while maintaining character consistency, achieving a stable balance between the two. We deconstruct the text-to-image model for ID customization, dividing it into an image generation branch and a character feature branch. The character feature branch is further decoupled into three modules: feature extraction, feature fusion, and feature integration. In feature extraction, we isolate five layers of identity-aware features from the fine-grained local feature extractor EvaCLIP \cite{sun2023eva}, which we term “mapping features.” In feature fusion, due to the one-to-one correspondence between mapping features and five groups of neural networks, we designate features outside the five-layer mapping features of EvaCLIP as “shift features.” We discovered that these two types of features essentially perform semantic compression of local features across network depths, forming an editable feature space. The combination of these features enables predictable editability variations. In feature integration, when the fused features interact with the image generation branch, we introduce a dynamic information fusion mechanism in cross-attention to further enhance editability, ultimately achieving state-of-the-art (SOTA) results in the IBench evaluation framework.

Additionally, most current models are based on the UNet architecture, with SD \cite{rombach2022high} and SDXL \cite{podell2023sdxl} serving as foundational models. In practical applications, such as story generation scenarios, there are typically two requirements: 1) the input of long prompts, and 2) higher aesthetic quality for the generated images. Consequently, image generation based on the DiT \cite{peebles2023scalable} architecture, such as Flux \cite{flux2024}, becomes the preferred choice. However, there is scarcely any literature discussing methods to enhance character editability while preserving character consistency in DiT-based image generation. To the best of our knowledge, we are the first to explore improving character editability within the DiT architecture.

Considering the current lack of unified datasets and metrics for evaluating both ID consistency and editability in the field of ID customization, we propose a configurable and modular automated evaluation framework. We designed multiple sets of evaluation character images, accounting for real IDs and generated IDs, and adapted them to various types of prompts, including editability-measuring prompts, short prompts, and more. For the first time, we comprehensively introduce a variety of editability verification metrics to quantitatively assess model performance.

In summary, our contributions are as follows:
\begin{enumerate}
    \item EditID is the first character customization method to address editability enhancement within the DiT architecture.
    \item We propose an editability enhancement approach under a training-free framework, utilizing mapping features, shift features, and dynamic ID integration across three modules. By decoupling the character feature branch, we achieve a highly editable text-to-image model while maintaining ID consistency.
    \item We introduce a unified ID consistency evaluation framework and, for the first time, propose multiple editability metrics. EditID demonstrates excellent performance on editability-focused prompts.
\end{enumerate}

\section{Related Works}
\subsection{ID Consistency Generation}
Leveraging the powerful generative capabilities of T2I diffusion models, numerous personalized generation methods have emerged. ID consistency generation is a specialized form of personalized generation that focuses on strong semantic facial features and has broad applications in the T2I domain. Early representative work, such as IP-Adapter-FaceID \cite{IPAdapterFaceID2024}, utilized facial embedding features extracted from face recognition models, combined with decoupled cross-attention feature integration, to maintain ID consistency. Photomaker \cite{li2024photomaker} encodes an arbitrary number of input ID images into a stacked ID embedding to preserve ID information. InstantID \cite{wang2024instantid} designs IdentityNet, integrating facial features, landmarks, and textual information through semantic and weak spatial constraints to guide image generation. FastComposer \cite{xiao2024fastcomposer} introduces cross-attention localization supervision, forcing the attention of the reference subject to focus on the correct regions of the target image, and proposes delayed subject conditioning during denoising steps to maintain identity consistency. PuLID \cite{guo2024pulid} combines a Lightning T2I branch with a standard diffusion branch, introducing contrastive alignment loss and precise ID loss to minimize interference with the original T2I model, ensuring strong ID fidelity.

\subsection{ID Editable Methods}
Currently, few text-to-image works in ID customization explicitly prioritize editability as a key improvement focus; most efforts still center on the dimension of ID consistency. Among the limited studies addressing editability, DreamIdentity \cite{chen2024dreamidentity} proposes the M2ID encoder, featuring ID-aware multi-scale features with multi-embedding projection, and introduces a novel self-augmented editability learning approach to enhance the model’s editing capabilities. PortraitBooth \cite{peng2024portraitbooth} leverages subject embedding features from face recognition models and incorporates emotion-aware cross-attention control to achieve diverse facial expressions in generated images, supporting text-based expression editing. ConsistentID \cite{huang2024consistentid} includes multimodal facial prompt generation and an ID preservation network. The facial generator integrates local facial features, facial feature descriptions, and overall facial descriptions to improve the precision of facial detail reconstruction, while the ID preservation network is optimized through a facial attention localization strategy, ensuring consistent identity across facial regions in line with the facial descriptions.

Current methods for character consistency and editability are predominantly based on the UNet architecture, such as SD and SDXL, and are mostly tuning-free solutions. These approaches typically produce images with average texture quality, offer poor support for long prompts, and lack flexibility, requiring pre-training on large-scale facial datasets.

\subsection{Training-Free Framework}
The training-free framework differs from fine-tuning and tuning-free approaches and has been widely applied in the fields of image and video generation. FreeU \cite{si2024freeu} introduces two key modulation factors: the backbone feature factor and the skip-connection feature scaling factor. The former amplifies the effect of feature maps in the backbone network to enhance the denoising process, while the latter enables a balanced adjustment during denoising, ensuring that images retain clarity without losing excessive detail information. FreePromptEditing \cite{liu2024towards} performs image editing by replacing self-attention maps in specific attention layers during the denoising process, allowing modifications to objects or object attributes in the image. Freelong \cite{lu2024freelong} combines the low-frequency components of global video features with the high-frequency components of local video segments, integrating diverse and high-quality spatiotemporal details from local videos while maintaining global consistency, thus improving the coherence and fidelity of long video generation.

To the best of our knowledge, we are the first to introduce the training-free framework to the domain of customized ID editable in text-to-image generation under the DiT architecture.

\section{Method}
\begin{figure*}[t]
  \centering
  \includegraphics[width=\textwidth]{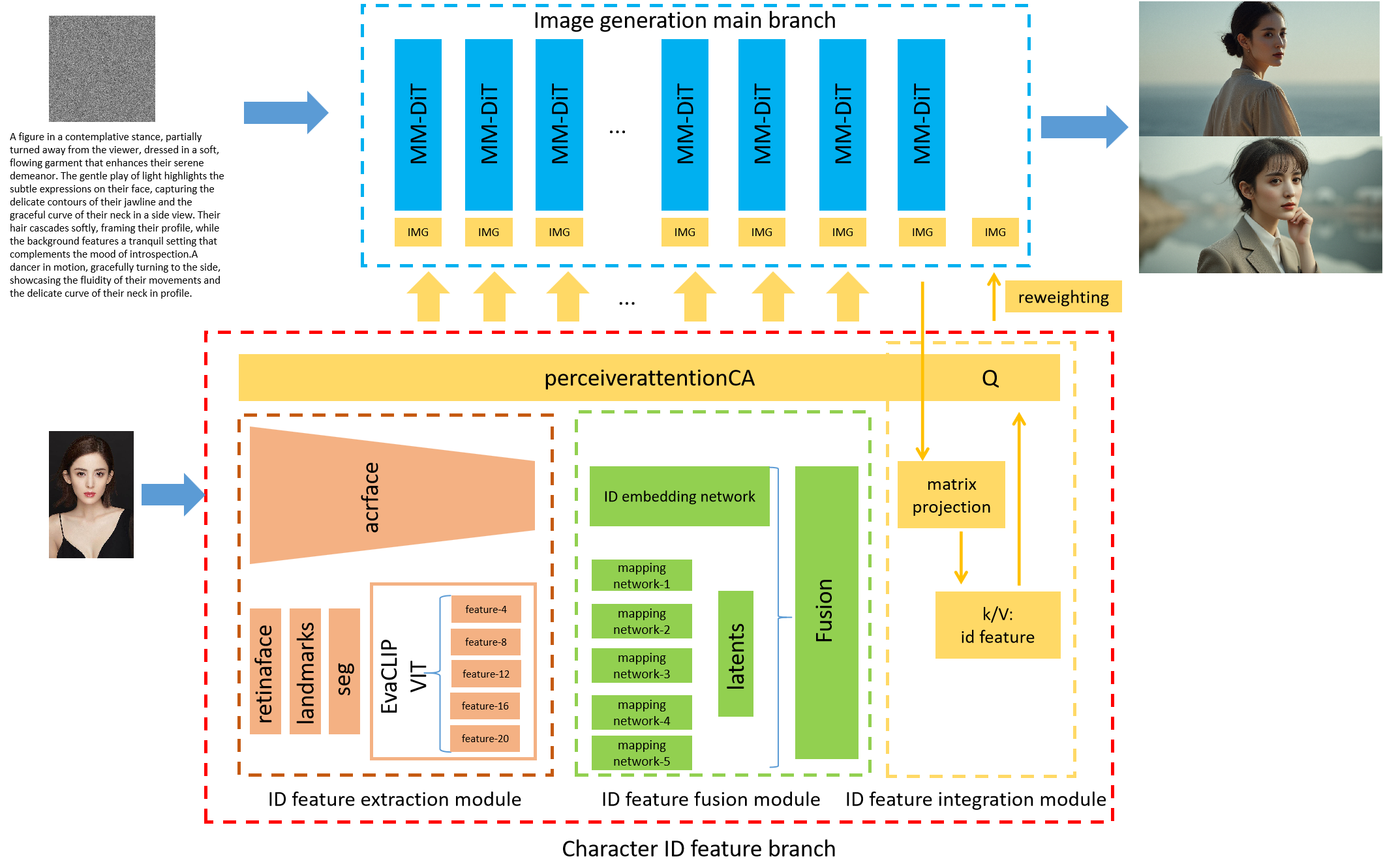}
  \caption{Overview of the EditID Framework. The upper half of the framework depicts the DiT-based image generation process. The lower half represents the character feature branch, which is divided into three parts. The first part is the ID feature extraction Module, responsible for extracting global and local features, generating mapping features. The second part is the ID feature fusion module, tasked with fusing the mapping features, producing shift features at this stage. The third part is the ID feature integration module, which implements the dynamic ID embedding mechanism design.}
  \label{fig:figure2}
\end{figure*}

The core of our approach is to prioritize the enhancement of character editability while ensuring character consistency. \cref{fig:figure2} illustrates the architecture of our method, which deconstructs the ID customization approach into an image generation main branch and a character feature branch. The feature branch is further decoupled into three modules: character feature extraction, feature fusion, and feature integration. This framework simultaneously identifies the sources of editability: 1) local ID features from the feature extraction module; 2) ID shift features from the feature fusion module; and 3) the embedding strength design for integrating ID information into the DiT generation backbone. Notably, the third aspect influences not only editability but also character consistency. Through the sophisticated combination and design of local ID features, we achieve optimal editability.

\subsection{Preliminary}
\subsubsection{DiT Flow Matching}
The core of diffusion models lies in achieving data generation through a progressive denoising process. Traditional diffusion models define the forward diffusion process as
\begin{equation}
  q(x_t|x_0) = \mathcal{N}(x_t; \alpha_t x_0, \sigma_t^2 I)
  \label{eq:diffusion}
\end{equation}
where $\alpha_t$ and $\sigma_t$ are noise scheduling coefficients, and $t \in [0, T]$ represents continuous time steps. Based on flow matching theory \cite{esser2024scaling}, the generation process can be modeled as an ordinary differential equation (ODE):
\begin{equation}
  dx = v_\theta(x_t, t, c) dt
  \label{eq:ode}
\end{equation}
where $v_\theta$ is the vector field to be learned, and $c$ is the conditional input (e.g., text prompts). Compared to traditional diffusion models, which rely on noise prediction targets, flow matching directly learns the transport mapping from the data distribution to the noise distribution. Its training objective can be expressed as:
\begin{equation}
  \mathcal{L}_{FM}(\theta) = \mathbb{E}_{t,q(x_0),p(x_1)} \left[ \|v_\theta(x_t,t,c) - (x_1 - x_0)\|^2 \right]
  \label{eq:fm_loss}
\end{equation}
where $x_t = (1 - t)x_0 + t x_1$ represents a linear interpolation path, and $t \in [0, 1]$.

In the DiT architecture, we replace the traditional UNet with Transformer modules, leveraging the self-attention mechanism to model global context. Given a conditional embedding sequence $c$, the vector field predictor in DiT can be decomposed as:
\begin{equation}
  v_\theta = \text{Proj}(\text{Attn}(Q,K,V))
  \label{eq:dit_attn}
\end{equation}
where $Q = x_t W_Q$, $K = c W_K$, and $V = c W_V$ are the query, key, and value matrices, respectively. $W$ denotes learnable projection matrices, and $\text{Attn}$ represents the multi-head attention mechanism. This architecture is particularly well-suited for long-text conditional generation, as its self-attention mechanism effectively captures long-range dependencies between prompts.

Currently, image generation models based on the DiT architecture and flow matching are primarily represented by SD3 \cite{esser2024scaling} and Flux \cite{flux2024}. This paper selects Flux as the foundational framework. The Flux framework builds upon DiT by introducing a regularized flow matching strategy, employing an improved noise scheduling function:
\begin{equation}
  \alpha_t = \cos^2(\pi t/2), \quad \sigma_t = \sin^2(\pi t/2)
  \label{eq:noise_schedule}
\end{equation}
This ensures a smooth transition from data to noise while maintaining numerical stability during the flow matching process.

\subsubsection{DiT-Based ID Consistency}
ID customization methods based on DiT are exceedingly rare, with most character ID customization approaches still relying on the UNet architecture of SD and SDXL. FluxCustomID \cite{FluxCustomID2024} employs ArcFace \cite{deng2019arcface} and CLIP \cite{hafner2021clip} to extract fine-grained ID features, which are ultimately embedded into the image generation branch via PerceiverResampler \cite{alayrac2022flamingo}. However, in practical testing, its performance across three dimensions—fidelity, task consistency, and editability—remains suboptimal. PuLID \cite{guo2024pulid} is currently the state-of-the-art model for ID customization. PuLID has introduced an experimental version based on Flux, addressing two significant challenges: 1) ID embedding interferes with the behavior of the original model, and PuLID aims to preserve the original T2I generation capability after ID insertion; 2) lack of fidelity. To tackle these, PuLID introduces a Lightning T2I branch alongside the standard diffusion denoising training branch, forming a contrastive pair. This pair shares the same prompts and initial latent variables, with one undergoing ID insertion and the other not. During the Lightning T2I process, features between the contrastive pair are semantically compared to guide the ID adapter in inserting information without disrupting the original model’s behavior. Additionally, ID embedding information is extracted and compared with real facial features to compute an ID loss, enhancing fidelity. PuLID’s character feature branch uses ArcFace to extract global features and, after detecting, aligning, and segmenting the face, employs EvaCLIP to extract local features. In this paper, we adopt PuLID as the baseline model, leveraging its overall design of ID loss and alignment loss. Within a training-free framework, we explore the patterns of editability variation by segmenting and designing the character feature branch into three modules.

\subsection{Feature Mapping}
\begin{figure}[t]
  \centering
  \includegraphics[width=\linewidth]{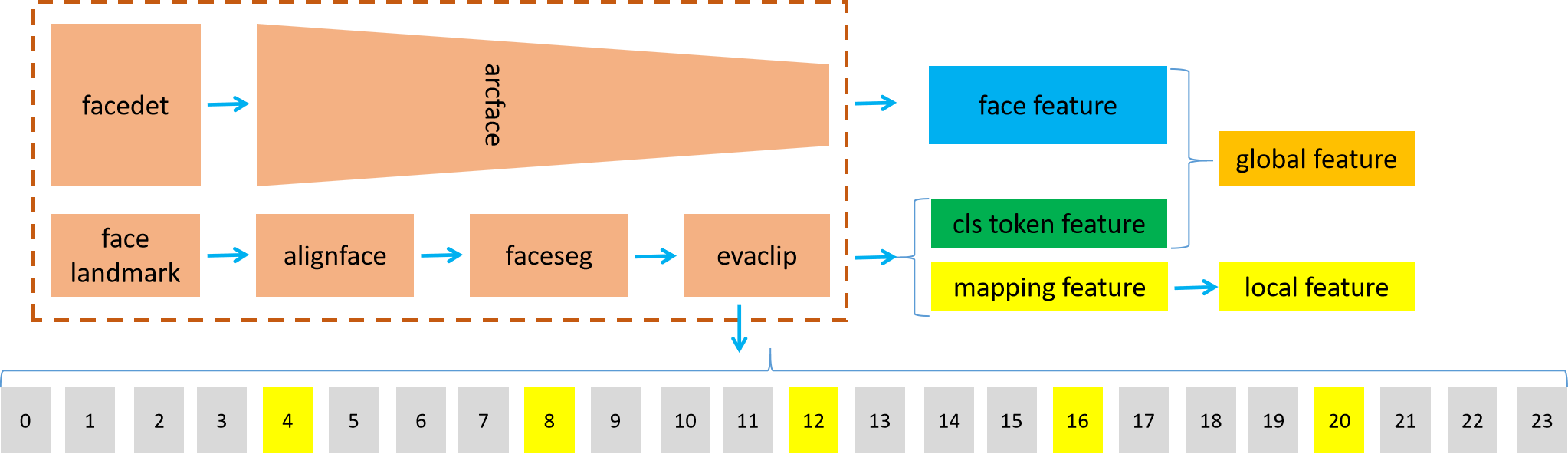}
  \caption{Detailed diagram of the character feature extraction module. The blue facial features and the dark green CLS token features from EvaCLIP together form the global features. The yellow features represent the mapping features, with layers 4, 8, 12, 16, and 20 in EvaCLIP being the original mapping features. Gray indicates the unselected features.}
  \label{fig:figure3}
\end{figure}

In the character feature extraction module, most approaches to character feature processing combine global and local features. We discovered that editability is embedded within local features. The feature extraction module of EditID consists of two branches. Branch 1: The character feature extractor employs SCRFD \cite{deng2021masked} for lightweight face detection, and the detected facial region is then processed by ArcFace for global feature extraction. Branch 2: RetinaFace is used for facial keypoint detection, identifying five keypoints, followed by frontal face alignment. After facial semantic segmentation, a more refined facial region is obtained, and EvaCLIP is then applied for fine-grained local feature extraction. From EvaCLIP’s 23-layer features, we select five identity-aware layers. Surprisingly, we found that the selection of these five identity-aware layers significantly enhances editability. We refer to these five identity-aware features as "mapping features." The global features are composed of the face features from Branch 1 and the CLS token features from Branch 2, while the mapping features directly correspond to local features. Ultimately, both global and local feature sets are output and fed into the feature fusion module. A detailed diagram of the module is shown in \cref{fig:figure3}.

\begin{figure*}[t]
  \centering
  \includegraphics[width=\textwidth]{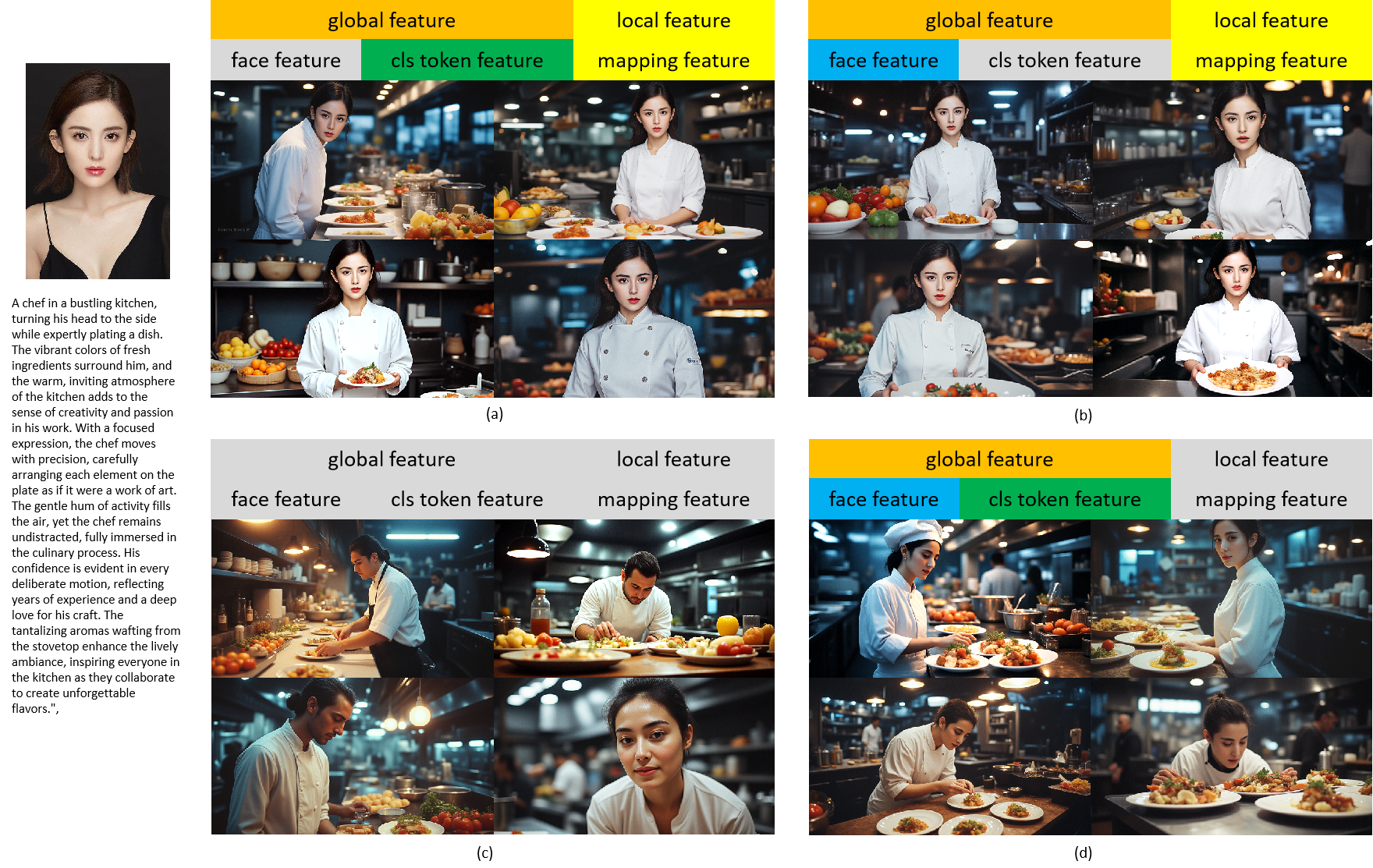}
  \caption{Impact of global and local features on generation editability. Gray indicates unselected features, set to zero.}
  \label{fig:figure4}
\end{figure*}

\begin{figure}[t]
  \centering
  \includegraphics[width=0.9\linewidth]{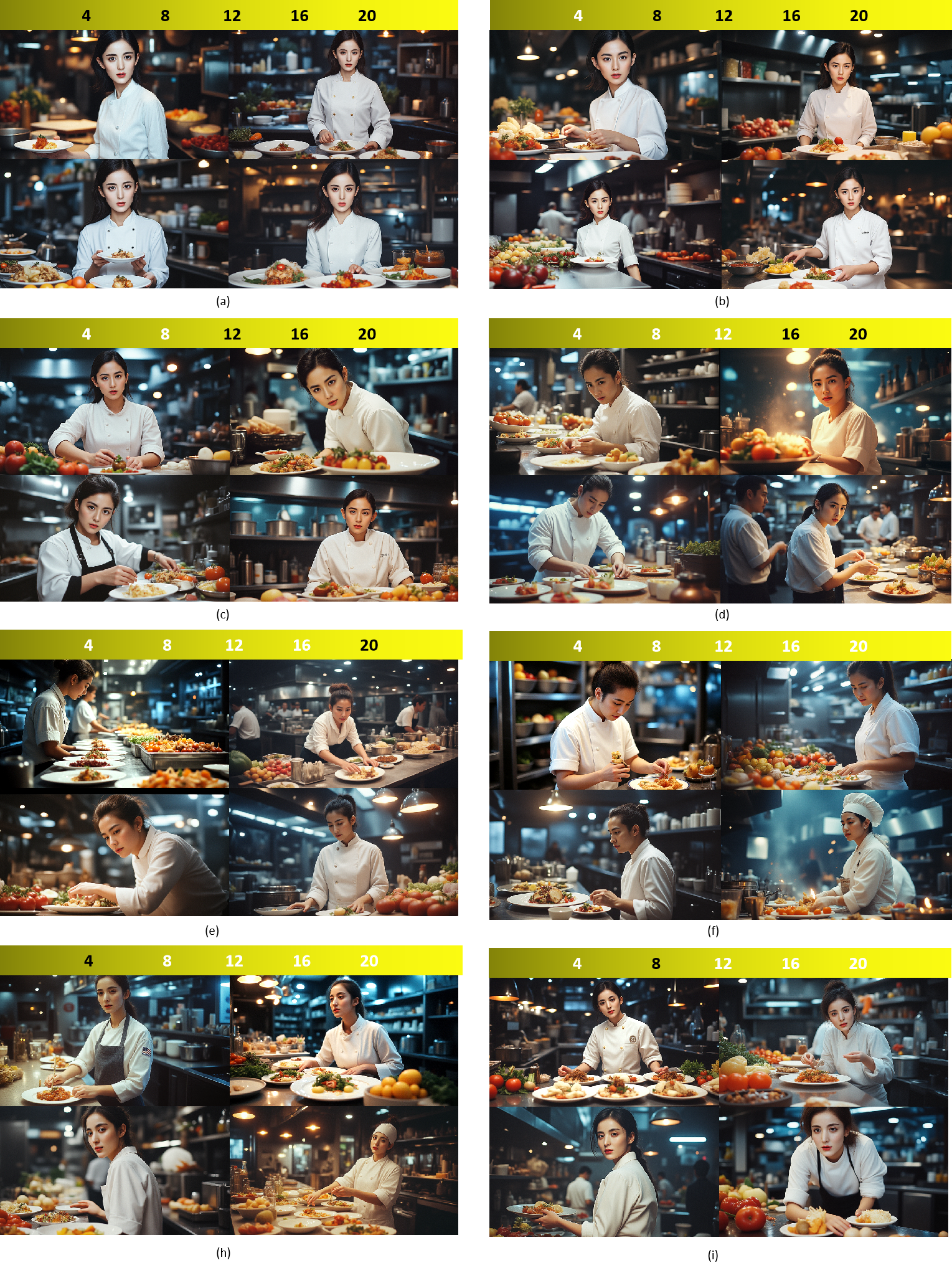}
  \caption{Impact of global and local features on generation editability. Gray indicates unselected features, set to zero.}
  \label{fig:figure5}
\end{figure}

We conducted further analysis on global and local features. When inputting prompts with noticeable action changes, as shown in \cref{fig:figure4}(a) and \cref{fig:figure4}(b), setting the facial features or CLS token features of the global features to zero revealed no significant changes in limb or facial orientation. However, \cref{fig:figure4}(b) exhibited higher character consistency than \cref{fig:figure4}(a), indicating that ArcFace extracts finer-grained facial features compared to CLIP, capturing distinctions more effectively. In \cref{fig:figure4}(c), when all features and local features were set to zero, the character feature branch became ineffective, degenerating into standard Flux image generation. This significantly improved editability but eliminated character consistency. In \cref{fig:figure4}(d), setting local features to zero still preserved good editability, though character consistency decreased. This led us to discover that global features predominantly control character consistency, while editability is concealed within local features. Global features tend to encode overall ID information of the face, such as facial structure, exhibiting high coupling and stability. In contrast, local features, through the identity-aware filtering of EvaCLIP, reduce 23 layers of features to 5, essentially achieving semantic compression across network depths. This process decouples features of different facial attributes at a fine-grained level, forming independently operable semantic units. We performed a more granular decomposition of the mapping features in EvaCLIP. As shown in \cref{fig:figure5}, from \cref{fig:figure5}(a) to \cref{fig:figure5}(f), we observed that as the mapping features were filtered, editability changed accordingly, but character consistency was also affected. An increase in editability corresponded to a decrease in character consistency. This prompted us to explore the optimal balance point between consistency and editability.

This module consists of two parts: global features and cross-layer local features (mapping features). Its core objective is to decouple the global features that control ID consistency from the local features that carry the freedom of editability. As shown in Branch 1, global features are jointly extracted by two-dimensional encoders:
\begin{equation}
  F_{\text{global}} = [\Psi_{\text{arcface}}(I); \Phi_{\text{CLS}}(I)] \in \mathbb{R}^{d_g}
\end{equation}
where $\Psi_{\text{arcface}}: \mathbb{R}^{H \times W \times 3} \rightarrow \mathbb{R}^{d_a}$ represents the ArcFace-based dense facial encoder, responsible for extracting deep semantic features related to facial ID; $\Phi_{\text{CLS}}: \mathbb{R}^{H \times W \times 3} \rightarrow \mathbb{R}^{d_c}$ denotes the CLS token feature from the EvaCLIP image encoder, capturing the overall visual context information of the character. The two are fused through a concatenation operation $[;]$ into a global feature vector of dimension $d_g = d_a + d_c$.

As shown in Branch 2, in the cross-layer local features, editable local features are obtained through hierarchical semantic compression:
\begin{equation}
  F_{\text{local}} = \{\phi_l(I)\}_{l \in L_{\text{map}}} \in \mathbb{R}^{5 \times d_l}
\end{equation}
where $L_{\text{map}} = \{l_1, l_2, \dots, l_5\}$ represents the five feature layers selected from the 23 layers of EvaCLIP, and $\phi_l(I)$ denotes the output feature map of the $l$-th layer. Through a cross-layer filtering mechanism, feature responses strongly correlated with facial attributes (e.g., expression, orientation) are isolated, forming independently manipulable semantic units.

\subsection{Feature Shift}
In the feature fusion module, global features are input into the ID embedding network, a neural network consisting of three linear layers, while local features are fed into the mapping network. The mapping network shares a similar structure to the ID embedding network, and both facilitate feature transformation. Our method is based on a training-free architecture. Therefore, during the tuning phase, the mapping features and the mapping network establish a one-to-one mapping relationship. When we replace the mapping features with shift features, a feature shift occurs between them and the mapping network. The mapping features consist of only five groups, selected as five identity-aware features from the 23 layers of EvaCLIP. This selection essentially achieves semantic compression across network depths: shallow layers capture compositional structure information, middle layers encode detailed geometric structures, and deep layers associate with high-level semantics. This hierarchical selection constructs an editable semantic space, where different layers correspond to facial editing dimensions of varying granularity. Visualizing the facial features of EvaCLIP provided us with guidance for selecting features. We found that the choice of shift features also significantly impacts editability, and we ultimately selected the feature combination of layers 4, 14, 16, 18, and 20.

\begin{figure}[t]
  \centering
  \includegraphics[width=\linewidth]{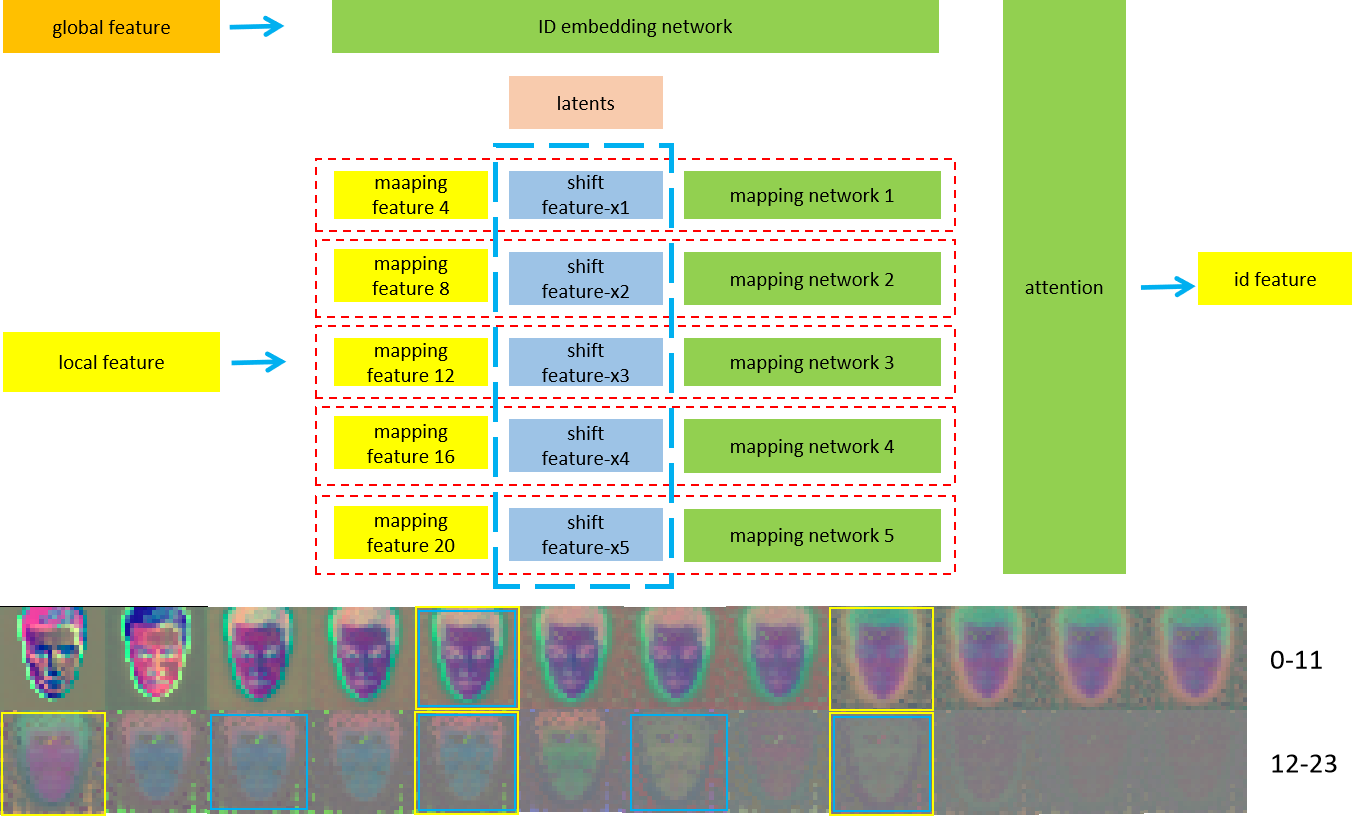}
  \caption{The upper half depicts the feature fusion module, where shift features are introduced. The lower half shows the visualization of facial features from the 23 layers of EvaCLIP, with yellow modules representing mapping features and blue boxes indicating the combination of shift features and mapping features.}
  \label{fig:feature_shift}
\end{figure}

This module fuses global and local features, introducing richer editability through mapping features and shift features. The formula is as follows:
\begin{equation}
  F_{\text{edit}} = \text{Attn}\big(\theta_{\text{ID}}(I), M_{\text{Map}}(F'_{\text{local}})\big) \in \mathbb{R}^{d_{\text{id}}},
\end{equation}
where $F'_{\text{local}} = [F_{\text{local}}(l); F_{\text{shift}}(l)]$ and $F_{\text{edit}}$ is the output combined feature, $\theta_{\text{ID}}(I)$ is derived from an ID embedding network composed of three linear layers, and $M_{\text{Map}}$ represents the mapping network. Here, $F'_{\text{local}}$ is the combination of mapping features and shift features, $F_{\text{local}}$ denotes the local features composed of mapping features, and $F_{\text{shift}}$ represents the shift features. The total number of mapping features and shift features satisfies:
\begin{equation}
  |L_{\text{map}}| + |L_{\text{shift}}| = 5,
\end{equation}
where $L_{\text{map}}$ and $L_{\text{shift}}$ denote the sets of mapping features and shift features, respectively.

\subsection{ID Feature Integration}
\begin{figure}[t]
  \centering
  \includegraphics[width=0.9\linewidth]{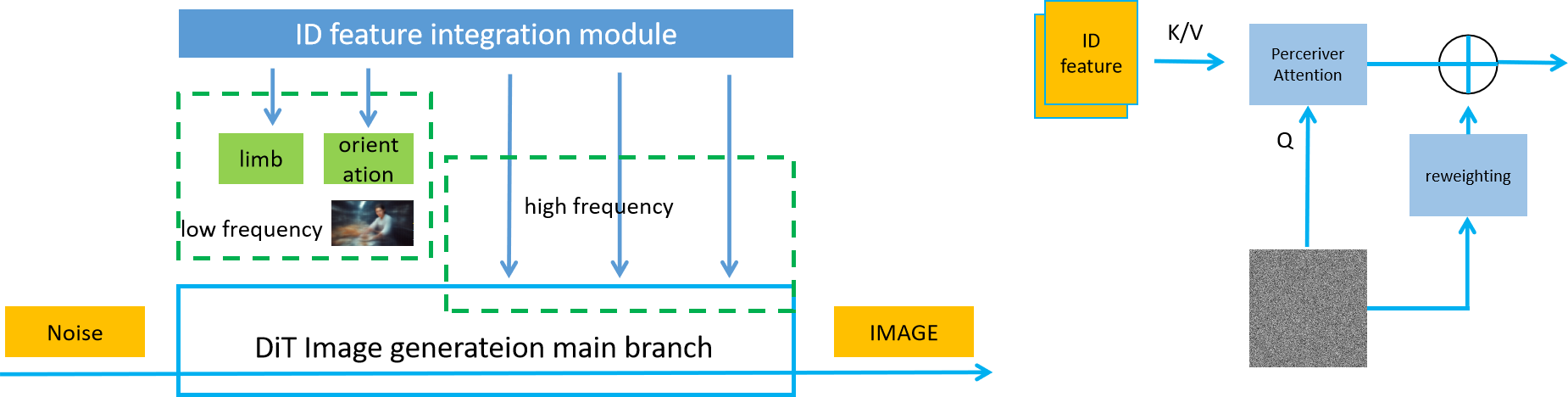}
  \caption{The left figure shows the dynamic ID information interaction between the ID feature integration module and image generation main branch, while the right figure illustrates the soft control of ID information within the ID feature integration module.}
  \label{fig:dynamic_id}
\end{figure}

After the feature fusion module, only one set of features is output to the ID feature integration module. This module needs to interact with the image generation main branch through PerceiverAttention. In Flux’s 19 dual-stream blocks and 38 single-stream blocks, we select 10 blocks for embedding ID information. We explored this in two dimensions. First, the early stages of image generation primarily involve low-frequency information, such as color and composition, while the middle and later stages involve high-frequency information, corresponding to the mid-to-late sampling steps that optimize details. Therefore, we adjusted the intensity of ID information embedding during the initial denoising phase, as shown in \cref{fig:dynamic_id}. However, excessively strong ID embedding can disrupt the balance of the noise distribution, preventing the model from correctly decoding low-frequency information. This blunt adjustment overlooks the progressive nature of the generation process, introducing an offset in the initial generation that undermines convergence. As a result, the generated images become darker overall, leading to losses in lighting and stability.

Secondly, we adopted a softer approach to ID strength control. In the feature integration module, the generated noise image serves as the Query, while the ID information acts as the Key and Value for weighting. During output, we perform reweighting on the Query to align it with the same dimensional size as the ID feature, followed by information supplementation through a residual connection, using concatenation for fusion. Reweighting can be implemented in various ways. To achieve dimensional transformation without excessively weakening the generated noise, we designed a transformation matrix. Considering the characteristics of information retention, we explored methods such as randn linear, DCT, and partial Fourier, ultimately adopting the randn linear approach. The editability of image generation primarily stems from the image side. However, text information is embedded within the noise image, which essentially serves as the starting point of the denoising process and contains latent semantic information. By compensating for the noise image, the semantic influence of the text can more smoothly permeate the image generation process that integrates ID embedding information. This is equivalent to introducing additional degrees of freedom in the latent space, enabling text-driven editability to be realized under the constraints of ID information without being overly restricted by the ID embedding.

Ultimately, by combining the mapping features of the ID feature extraction module, the shift features of the ID fusion module, and the soft ID strength control mechanism of the ID integration module, we achieved excellent editability while preserving ID consistency..

\section{IBench}
\label{sec:ibench}

To address the lack and inadequacy of evaluation metrics for character consistency and editability in the field of personalized character image generation, and to quantify the improvement effects under a training-free architecture, we propose a configurable and modular automated evaluation framework, IBench. This framework comprehensively introduces and implements a variety of editability verification metrics.

\subsection{Dataset}
\label{subsec:dataset}

The evaluation data of IBench consists of two parts: prompts and evaluation images. The evaluation images are divided into three groups: Unsplash, ChineseID, and GenerateID. The Unsplash group includes 49 images, covering a variety of skin tones, significant variations in character angles, and instances of facial occlusion. ChineseID comprises 100 images of Chinese individuals collected from the internet, including well-known figures from fields such as film and sports, representing diversity in gender, age, and multiple angles. GenerateID consists of 100 ID images generated by a text-to-image model, featuring refined facial features, diverse poses, accessories, and hairstyles, as well as characters under light and shadow rendering, ensuring both aesthetic quality and prominent ID characteristics.

The prompts in IBench are categorized into three dimensions: short prompts, action prompts for editability, and manually collected action-story prompts. Short prompts, designed to be compatible with the evaluation of the UNet architecture, are largely sourced from mainstream evaluation reports. There are 20 groups of short prompts, encompassing various style attributes such as action, view, style, and complex. Editable long prompts are sourced from the augmented prompts in VBench~\cite{huang2024vbench}, with 41 groups selected, including long descriptions of character actions and scene stories, serving as a key focus of this evaluation. Manually collected prompts consist of 80 groups of prompts collected from text-to-image users, manually curated to include rich action descriptions with story elements.

In IBench, we pair Unsplash with short prompts as one group, ChineseID with editable long prompts as another group, and GenerateID with manually collected prompts as a third group. However, in practice, these images and prompts can be cross-validated. The prompts and images will be showcased in detail in the Appendix.

\subsection{Evaluation Metrics}
\label{subsec:evaluation_metrics}

We designed the metrics from three dimensions: consistency, editability, and the T2I general evaluation dimension. In the T2I general evaluation, the focus is primarily on the aesthetic attributes of the images. In the consistency dimension, the evaluation centers on the similarity of character ID generation and prompt-following consistency. In the editability dimension, we propose multiple innovative measurement indicators to assess the editability of character IDs.

\subsubsection{T2I general evaluation dimension}

\noindent \textbf{FID}: By comparing the distribution differences between \texttt{imagewithid} (generated images with ID information) and \texttt{imagewithoutid} (generated images without ID information) in the feature space of a pre-trained InceptionV3 model, we quantify the similarity between the two distributions. A lower value indicates higher similarity, suggesting that ID insertion does not affect image generation.

\noindent \textbf{Aesthetic}: The LAION aesthetic predictor is used to evaluate the aesthetic quality score of \texttt{imagewithid} images. It reflects dimensions such as the harmony and richness of layout and color, as well as the realism and naturalness of the images.

\noindent \textbf{Imaging Quality}: This assesses distortions (e.g., overexposure, noise, blur) present in \texttt{imagewithid} images, using the MUSIQ~\cite{ke2021musiq} image quality predictor trained on the SPAQ dataset for evaluation.

\subsubsection{Consistency Dimension}
\noindent \textbf{Facesim}: This calculates the facial similarity between the ID image and  \texttt{imagewithid}. We use SCRFD from InsightFace to detect facial regions and ArcFace to extract facial feature vectors, then compute the cosine similarity to measure the similarity of the generated facial regions.

\noindent \textbf{ClipT}: This computes the cosine similarity between the CLIP text encoding of the input prompt and the CLIP image encoding features of \texttt{imagewithid}. It is used to evaluate the ability of the generated images to follow changes in the prompt.

\noindent \textbf{ClipI}: This calculates the cosine similarity between the CLIP image encodings of \texttt{imagewithid} and \texttt{imagewithoutid}. It measures the similarity between the two images before and after ID insertion. A higher ClipI score indicates that the modifications to image elements after ID insertion cause less interference compared to the original model’s generation.

\noindent \textbf{Dino~\cite{zhang2022dino}}: This computes the cosine similarity between the DINO image encodings of the ID image and \texttt{imagewithid}. DINO features are more fine-grained and can be used to measure the changes in the generated image relative to the ID image.

\noindent \textbf{Fgis}: Corresponding to the DINO metric, this calculates the cosine similarity of the DINO image encodings for the facial regions of the ID image and \texttt{imagewithid}. Facial detection is performed using MTCNN, enabling fine-grained measurement of similarity in the facial regions.

\subsubsection{Editability Dimension}
\noindent \textbf{Posediv}: This calculates the differences in Euler angles (yaw, pitch, and roll) of the facial regions between the ID image and \texttt{imagewithid}. Facial detection is performed using MTCNN~\cite{yin2017multi}, and Euler angles are extracted using Hopenet~\cite{doosti2020hope}. This metric is used to assess the editability of the facial regions.

\noindent \textbf{Landmarkdiff}: This computes the difference in the average Euclidean distance of five normalized key points between the facial regions of the ID image and \texttt{imagewithid}. Facial detection is performed using MTCNN, and normalization is based on the maximum diagonal length of the bounding rectangle of the five points.

\noindent \textbf{Exprdiv}: This calculates the proportion of expression changes in the facial regions between the ID image and \texttt{imagewithid}. Facial detection is performed using MTCNN, and the expression classification model, based on VGG19, categorizes expressions into seven classes: Angry, Disgust, Fear, Happy, Sad, Surprise, and Neutral. The metric measures the proportion of expression changes in the ID before and after insertion.

\section{Experiments}
\label{sec:experiments}

\subsection{Setting}
\label{subsec:setting}

We use the Flux version of PuLID as the base model. For the Flux model, the sampling steps are set to 20, with a guidance scale of 3.5, a CFG scale of 1, and the Euler sampler is employed. For the character feature branch, we adopt Antelopev2 as the face recognition model and EVA-CLIP as the CLIP image encoder. Our final combination of mapping features and shift features consists of five groups [4, 14, 16, 18, 20], and we implement residual dynamic ID information embedding in the ID integration module, using concatenation as the fusion method. All experiments are conducted on four NVIDIA A100 GPUs, with the inference framework being ComfyUI.

\subsection{Qualitative Comparison}
\label{subsec:qualitative_comparison}

\begin{figure*}[t]
    \centering
    \includegraphics[width=\textwidth]{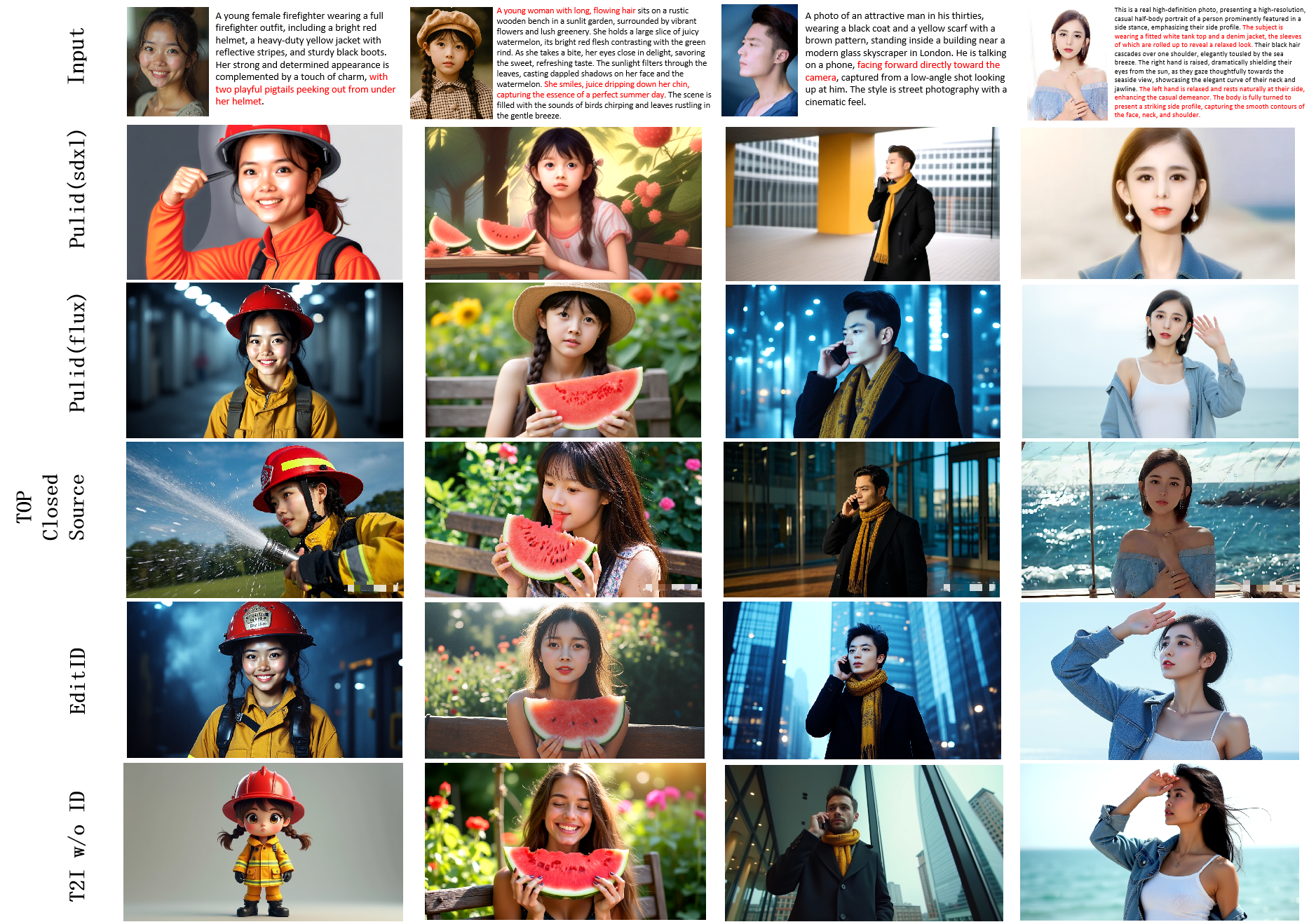}
    \caption{Qualitative Comparison: T2I w/o ID represents the output of Flux T2I without ID insertion. EditID achieves higher editability while maintaining ID consistency. It can accurately perform edits such as hairstyle and accessory changes (first column), age adjustments (second column), and facial and limb edits (third and fourth columns).}
    \label{fig:qualitative_comparison}
\end{figure*}

We adopt the SDXL version of InstantID, PuLID, and the Flux version of PuLID as the comparative model group, where the base models for InstantID and PuLID SDXL are \texttt{sdxl\_base\_1.0}. As shown in \cref{fig:qualitative_comparison}, EditID takes long text prompts as input and, while maintaining model consistency, achieves better editability compared to the nearly synchronous copy-paste face insertion of PuLID Flux. In the first column, when adding “with two playful pigtails peeking out from under her helmet”, EditID successfully alters the hairstyle. Compared to Flux T2I, the ID embedding enhances the ability to align the scene synchronously. In the second column, when adding “A young woman with long, flowing hair”, EditID enables age changes; however, the expression changes are less pronounced compared to the generation by Flux T2I. In the third column, with a side-profile ID image input and the addition of "facing forward directly toward the camera," EditID achieves facial rotation and completes the face when turned from a side profile to a front view. In contrast, PuLID Flux can hardly rotate the face. In the fourth column, with a front-facing ID image input and the addition of “The subject is wearing a fitted white tank top and a denim jacket, the sleeves of which are rolled up to reveal a relaxed look", EditID realizes a front-to-side profile transition while maintaining alignment with Flux T2I. The ID generation quality and detail handling of the PuLID SDXL version are generally poor, with low fidelity, possibly due to the limited generation capability of the SDXL base model. The character consistency is also inferior to the Flux version, and the generated images exhibit a weak non-realistic stylistic attribute. Top closed-source text-to-image models offer good richness in generation, producing scene associations not present in the prompts, but their fidelity and character consistency are both poor.

We primarily focus on the performance of EditID with long prompts. In \cref{tab:chineseid_evaluation}, using the evaluation combination of ChineseID with editable long prompts, EditID performs well across the three conventional aesthetic metrics: FID, Aesthetic, and Image Quality. In the Facesim metric, which measures consistency, EditID shows only a slight decrease compared to the Flux version of PuLID. However, as shown in \cref{fig:qualitative_comparison}, the Flux version of PuLID exhibits excessively strong consistency, even resulting in a copy-paste replication of the input ID face in the output, which significantly limits the applicability of text-image consistency generation. Facial and limb features need to exhibit different variations across various scenes. In the ClipI metric, it is evident that the ID insertion in EditID does not strongly interfere with the original generation capability. The ClipT metric also demonstrates good text-following ability, while Dino and Fgis, which are finer-grained consistency evaluation metrics, show significant improvements. For the most critical editability metrics, EditID achieves a total improvement of 5 points in the three Euler angles of Posediv compared to the Flux version of PuLID, and it also shows a substantial increase in Landmarkdiff. The Facesim metric decreases by only 2 points, indicating that EditID sacrifices only a slight degree of similarity while delivering excellent editability. In fact, the overly strong character consistency constraint in PuLID Flux suggests that releasing excessive consistency in exchange for enhanced editability is a highly prudent choice. Compared to the SDXL version of PuLID, while editability is high, it sacrifices too much character consistency.

\begin{table*}[t]
    \centering
    \makeatletter
    \def\@makecaption#1#2{%
        \vskip\abovecaptionskip
        \centering 
        \small #1: #2\par
        \vskip\belowcaptionskip
    }
    \makeatother
    \caption{Evaluation metric results from IBench on ChineseID with editable long prompts}
    \label{tab:chineseid_evaluation}
    \begin{tabular}{lccccccccc}
        \toprule
        Model & FID & Aesthetic & Image Quality & \multicolumn{3}{c}{Posediv} & Landmarkdiff & Exprdiv \\
        \cmidrule(lr){5-7}
        & & & & Yaw & Pitch & Roll & & \\
        \midrule
        InstantID & 9.780 & 0.585 & 0.394 & 12.62 & 7.545 & 5.266 & 0.044 & 0.647 \\
        PuLID (SDXL) & 11.28 & 0.675 & 0.502 & 19.48 & 6.187 & 12.69 & 0.099 & 0.593 \\
        PuLID (Flux) & 14.59 & 0.681 & 0.431 & 9.298 & 5.872 & 9.473 & 0.070 & 0.562 \\
        EditID & 13.52 & 0.683 & 0.454 & 11.81 & 6.722 & 10.60 & 0.082 & 0.554 \\
        \midrule
        & Facesim & ClipI & ClipT & Dino & Fgis & & & \\
        \cmidrule(lr){2-6}
        InstantID & 0.642 & 0.613 & 0.185 & 0.301 & 0.449 & & & \\
        PuLID (SDXL) & 0.399 & 0.768 & 0.248 & 0.129 & 0.353 & & & \\
        PuLID (Flux) & 0.735 & 0.757 & 0.243 & 0.178 & 0.501 & & & \\
        EditID & 0.714 & 0.769 & 0.249 & 0.162 & 0.459 & & & \\
        \bottomrule
    \end{tabular}
\end{table*}

\begin{table*}[t]
    \centering
    \makeatletter
    \def\@makecaption#1#2{%
        \vskip\abovecaptionskip
        \centering 
        \small #1: #2\par
        \vskip\belowcaptionskip
    }
    \makeatother
    \caption{Evaluation metric results from IBench on GenerateID with manually collected prompts}
    \label{tab:generateid_evaluation}
    \begin{tabular}{lccccccccc}
        \toprule
        Model & FID & Aesthetic & Image Quality & \multicolumn{3}{c}{Posediv} & Landmarkdiff & Exprdiv \\
        \cmidrule(lr){5-7}
        & & & & Yaw & Pitch & Roll & & \\
        \midrule
        InstantID & 12.87 & 0.613 & 0.403 & 13.31 & 7.994 & 5.977 & 0.046 & 0.662 \\
        PuLID (SDXL) & 8.514 & 0.663 & 0.530 & 18.61 & 7.093 & 10.13 & 0.087 & 0.541 \\
        PuLID (Flux) & 20.31 & 0.684 & 0.452 & 7.620 & 5.647 & 8.445 & 0.064 & 0.482 \\
        EditID & 19.28 & 0.682 & 0.464 & 9.795 & 6.881 & 9.683 & 0.075 & 0.485 \\
        \midrule
        & Facesim & ClipI & ClipT & Dino & Fgis & & & \\
        \cmidrule(lr){2-6}
        InstantID & 0.646 & 0.607 & 0.178 & 0.301 & 0.375 & & & \\
        PuLID (SDXL) & 0.384 & 0.768 & 0.252 & 0.189 & 0.390 & & & \\
        PuLID (Flux) & 0.724 & 0.729 & 0.231 & 0.286 & 0.579 & & & \\
        EditID & 0.701 & 0.739 & 0.238 & 0.262 & 0.475 & & & \\
        \bottomrule
    \end{tabular}
\end{table*}

We observed that the performance on ChineseID with editable long prompts and GenerateID with manually collected prompts is similar. Therefore, we will focus on the performance of ChineseID editable long prompts in subsequent analysis. Additional metrics can be found in Section 4 of the appendix.

\subsection{Ablation Study}
In our training-free framework, it is necessary to evaluate the impact of changes and combinations of multiple metrics. The following experimental groups primarily focus on similarity and editability metrics.

\subsubsection{Combination of Mapping and Shift Features}
In Section 3, we qualitatively analyzed the sources of editability variations from mapping features and shift features. The multi-level, fine-grained local semantic features introduced by EvaCLIP are the source of editability. Below, we quantitatively discuss this in two groups. In Table 3, we mainly examine the first two feature selections. In fact, the choice of the first feature is critical. As a shallow feature, it contains rich editable semantic information. In the first four groups, the first feature is the fourth-layer feature of EvaCLIP, and the second feature is the 8th, 12th, and 16th layers, respectively. It is evident that as semantic information weakens, Facesim decreases, but editability significantly increases, showing a strong inverse relationship. Comparing the third to sixth groups, where the first feature is replaced, we observe a decrease in Facesim but a greater increase in editability. This indicates that feature offsets at the same level (shallow layers) can provide substantial editability. Using IBench, we further dissect the combination information at a finer granularity.

\begin{table*}[t]
\centering
\makeatletter
\def\@makecaption#1#2{%
    \vskip\abovecaptionskip
    \centering 
    \small #1: #2\par
    \vskip\belowcaptionskip
}
\makeatother
\caption{Quantitative comparison of shift features on ChineseID with editable long prompts. "Features" represents a list of 5-layer features, and "-" indicates that the feature at that layer is set to 0.}
\begin{tabular}{lccccccc}
\toprule
Features & Facesim & ClipT & \multicolumn{3}{c}{Posediv} & Landmarkdiff & Exprdiv \\
\cmidrule(lr){4-6}
 & & & Raw & Pitch & Roll & & \\
\midrule
{[4,-,-,-,-]} & 0.512 & 0.262 & 27.46 & 11.28 & 13.17 & 0.115 & 0.621 \\
{[4,8,-,-,-]} & 0.635 & 0.256 & 17.59 & 8.288 & 11.46 & 0.096 & 0.589 \\
{[4,12,-,-,-]} & 0.617 & 0.258 & 19.25 & 9.033 & 11.89 & 0.102 & 0.591 \\
{[4,16,-,-,-]} & 0.591 & 0.259 & 23.10 & 9.689 & 12.41 & 0.108 & 0.581 \\
{[0,12,-,-,-]} & 0.601 & 0.261 & 20.51 & 9.343 & 12.18 & 0.106 & 0.593 \\
{[0,16,-,-,-]} & 0.574 & 0.262 & 24.75 & 10.41 & 13.26 & 0.112 & 0.585 \\
\bottomrule
\end{tabular}
\label{tab:shift_features}
\end{table*}

In Table 4, after selecting the shallow-layer features, we further examine the mid-layer and deep-layer features. Selecting only shallow-layer features significantly improves editability but results in greater loss of character consistency. When inputting rich long prompts, the generated images closely resemble ID-free text-to-image (T2I) results. We further screen mid-layer and deep-layer features, as deep-layer features provide richer details, contributing to improved image fidelity. We plotted the differences in Facesim and PoseDiv (raw/pitch/roll) values between our base model PuLID (Flux) and our proposed model in Fig 9. We observed that as the feature groups are adjusted, Facesim and PoseDiv exhibit a linear relationship. We selected the most cost-effective combination from the curve, ensuring high-level consistency while enhancing editability.




\begin{table*}[t]
\centering
\makeatletter
\def\@makecaption#1#2{%
    \vskip\abovecaptionskip
    \centering 
    \small #1: #2\par
    \vskip\belowcaptionskip
}
\makeatother
\caption{Quantitative comparison of feature combinations on ChineseID with editable long prompts. "Features" represents a list of 5-layer features, and "-" indicates that the feature at that layer is set to 0.}
\begin{tabular}{lccccccc}
\toprule
Features & Facesim & ClipT & \multicolumn{3}{c}{Posediv} & Landmarkdiff & Exprdiv \\
\cmidrule(lr){4-6}
 & & & Raw & Pitch & Roll & & \\
\midrule
{[4,12,18,20,-]} & 0.689 & 0.253 & 12.73 & 6.858 & 10.42 & 0.085 & 0.566 \\
{[4,14,18,20,-]} & 0.684 & 0.254 & 13.28 & 6.973 & 10.76 & 0.087 & 0.565 \\
{[4,16,18,20,-]} & 0.668 & 0.255 & 14.46 & 7.328 & 11.10 & 0.090 & 0.562 \\
{[4,12,16,18,20]} & 0.713 & 0.247 & 11.24 & 6.610 & 10.26 & 0.081 & 0.561 \\
{[4,14,16,18,20]} & 0.714 & 0.247 & 11.81 & 6.722 & 10.60 & 0.082 & 0.554 \\
{[4,16,16,18,20]} & 0.701 & 0.249 & 12.77 & 7.029 & 10.92 & 0.086 & 0.555 \\
\bottomrule
\end{tabular}
\label{tab:feature_combinations}
\end{table*}

\begin{figure}[t]
\centering
\includegraphics[width=\linewidth]{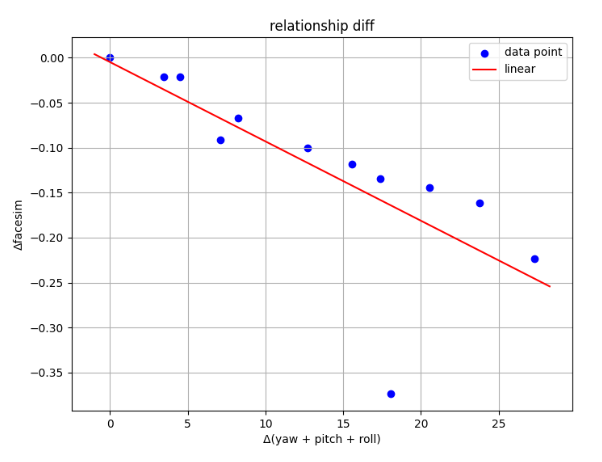}
\caption{Variation curves of the differences in raw/pitch/roll values from PoseDiv and Facesim compared to the corresponding values of PuLID (Flux).}
\label{fig:pose_div_facesim}
\end{figure}

\subsubsection{Shift Strategy}
\begin{figure}[t]
\centering
\includegraphics[width=\linewidth]{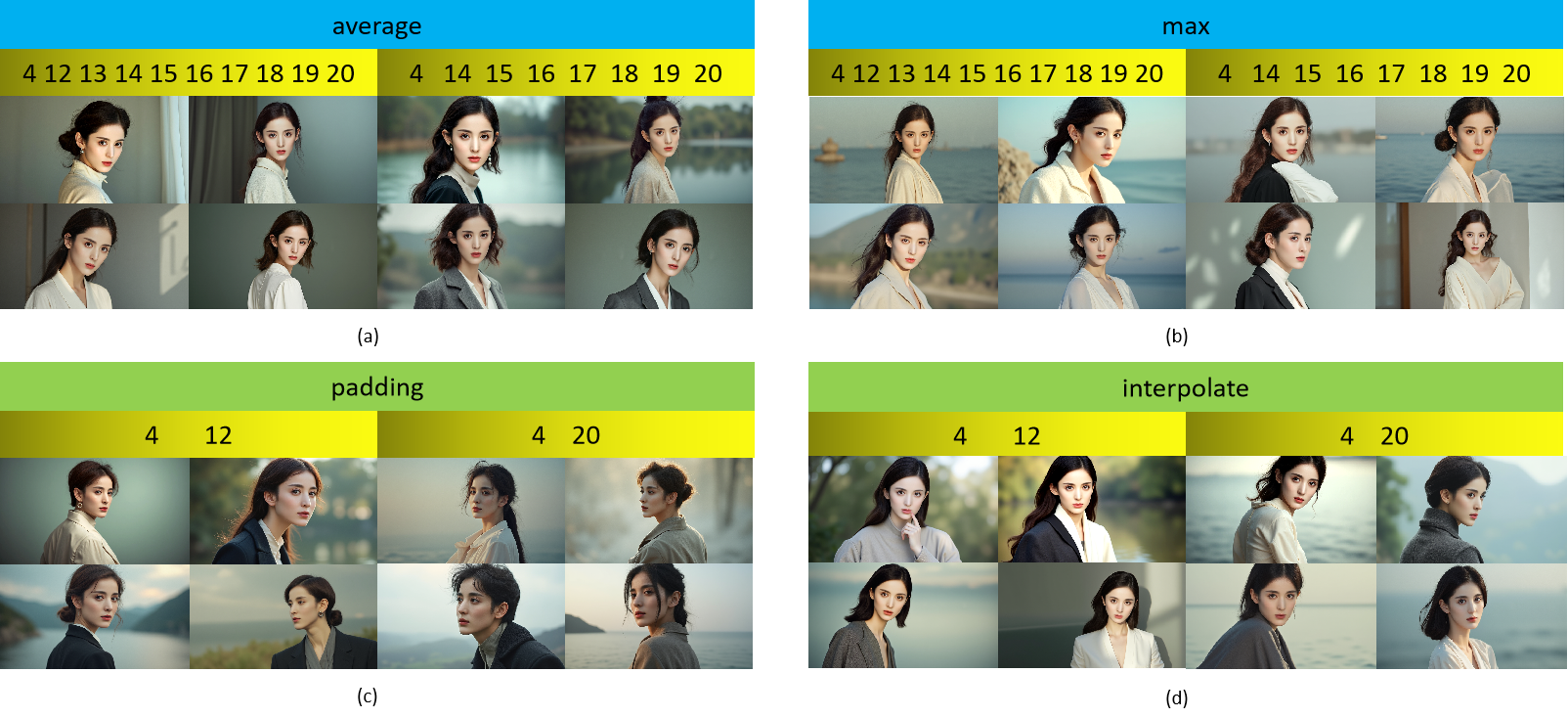}
\caption{Feature shift strategies, where the first row represents strategies for handling more than five sets of features, and the second row represents strategies for handling fewer than five sets of features.}
\label{fig:shift_strategies}
\end{figure}

For the combination of mapping features and shift features, the feature fusion module selects five sets of features to enter the mapping network, ultimately outputting ID features. These features are then integrated into the main image generation branch through the ID dynamic embedding mechanism in the ID integration module. For the design of these five sets of features, multiple approaches can be employed. When fewer than five sets of features are available, strategies other than zero-padding, such as interpolation, can be used. When more than five sets of features are available, strategies like average fusion can be applied. In \cref{fig:shift_strategies}, we qualitatively analyze two strategies for handling more than five sets of features (average and max) and two strategies for handling fewer than five sets of features (padding and interpolate). In Fig \cref{fig:shift_strategies}(a)(b), the max strategy results in images with higher sharpness and stronger lighting effects. In \cref{fig:shift_strategies}(c)(d), the interpolate strategy produces lower image quality. Selecting features from 23 layers yields better results than modifying features, which is why we prefer the combination of mapping features and shift features. The average method, based on the mean shift of existing features, also performs well.

\subsubsection{Editability in ID Integration Module}
\begin{figure}[t]
\centering
\includegraphics[width=\linewidth]{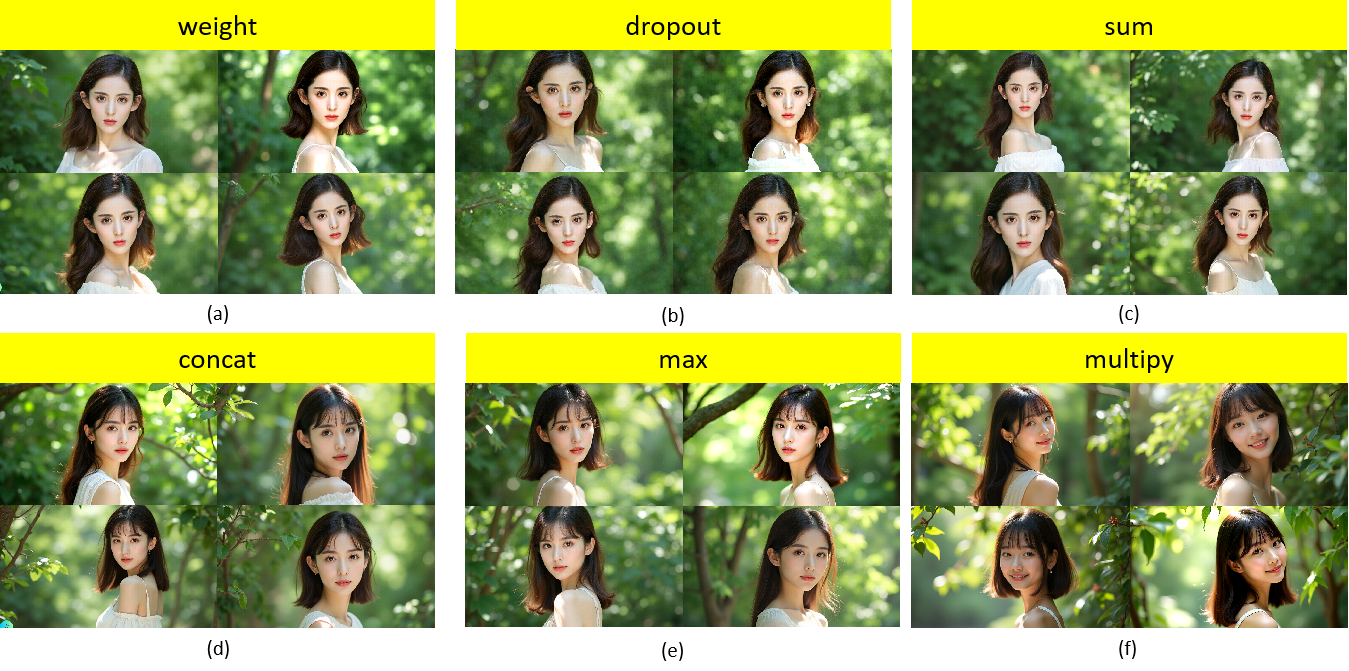}
\caption{Effect diagram of different feature fusion methods after reweighting.}
\label{fig:fusion_methods}
\end{figure}

The dynamic ID integration design in the ID integration module is also a significant source of editability. In this module, we primarily considered two dimensions: reweighting and feature fusion methods. Reweighting ensures dimensional consistency with the ID embedding features without compromising the noise features, while the fusion method appropriately compensates the features after ID integration back into the sampled noise features, enhancing editability in the text dimension. We conducted a qualitative analysis in \cref{fig:fusion_methods} to examine the impact of the randn linear approach in reweighting and various fusion methods. The fusion methods include: Weight: Assigning different fusion weights to the two feature sets; Dropout: Randomly masking features after reweighting to reduce information redundancy; Concat: Concatenating the two feature sets and then computing their mean for fusion; Sum: Directly summing the two feature sets; Multiply: Multiplying the two feature sets; Max: Taking the maximum of the two feature sets. We observed that in \cref{fig:fusion_methods}(a)(b)(c), the ID still exhibited a strong binding effect, with no significant changes in facial orientation, though the image fidelity decreased considerably. In \cref{fig:fusion_methods}(d)(e)(f), ID consistency declined, but editability gradually increased. Ultimately, we selected the concat fusion method from \cref{fig:fusion_methods}(d), combining it with mapping features and shift features to achieve a high level of consistency.

\section{Conclusion}
This paper proposes EditID, a training-free ID customization method for text to image generation. We are the first to explore enhancing editability within the DiT architecture, achieving state-of-the-art performance with long prompts. Taking the PuLID model as an example, we deconstruct it into a character feature branch and an image generation main branch, further decoupling the character feature branch into three major modules: feature extraction, feature fusion, and ID integration. We analyze the sources of editability from the combination of mapping features and shift features, as well as dynamic ID integration, thereby improving the editability of character customization. Our approach requires no training, demonstrating its potential for flexible and efficient character-customized image generation. Moreover, this training-free framework can be adapted to enhance any character ID customization generation algorithm equipped with a character feature branch. In future work, we will continue to explore and investigate the dynamic ID integration module with the introduction of a training mode. We believe that dynamic ID integration holds great vitality, but it still requires the design of loss functions incorporating richer multi-angle facial information to further achieve simultaneous improvements in both character consistency and editability.

{\small
\bibliographystyle{ieee_fullname}
\bibliography{egbib}
}
\newif\ifarxiv
\arxivtrue  
\ifarxiv \clearpage \appendix \input{appendix} \fi

\end{document}

%% file: appendix.tex
\section{Appendix}
\label{sec:appendix}
The structure of the supplementary material is as follows:

\subsection{More Details about IBench }
\label{subsec:More Details about IBench }
\subsubsection{Images}
\begin{figure}[t]
    \centering
    \includegraphics[width=\linewidth]{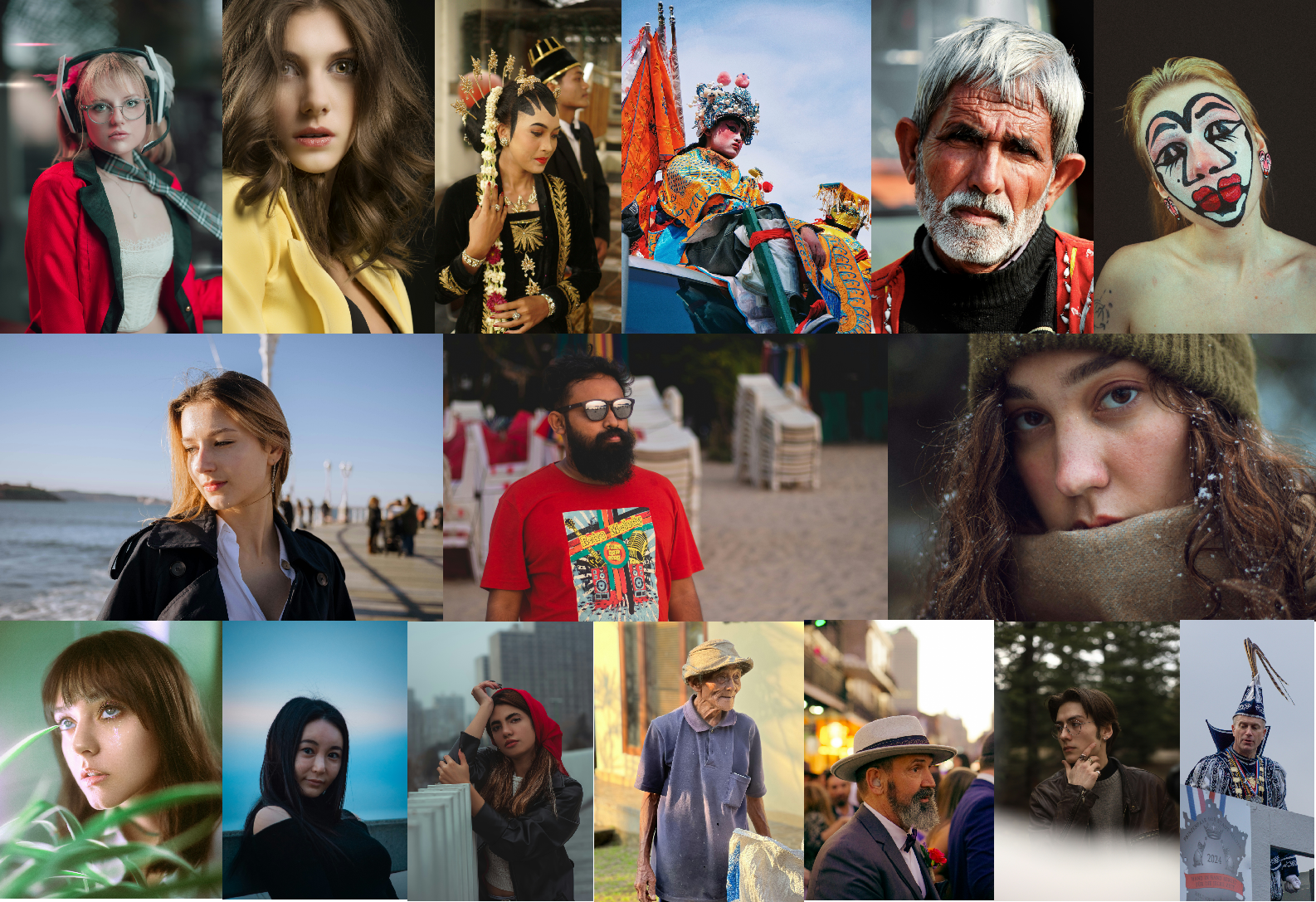}
    \caption{Examples of partial character ID images from Unsplash}
    \label{fig:unsplash}
\end{figure}

\begin{figure}[t]
    \centering
    \includegraphics[width=\linewidth]{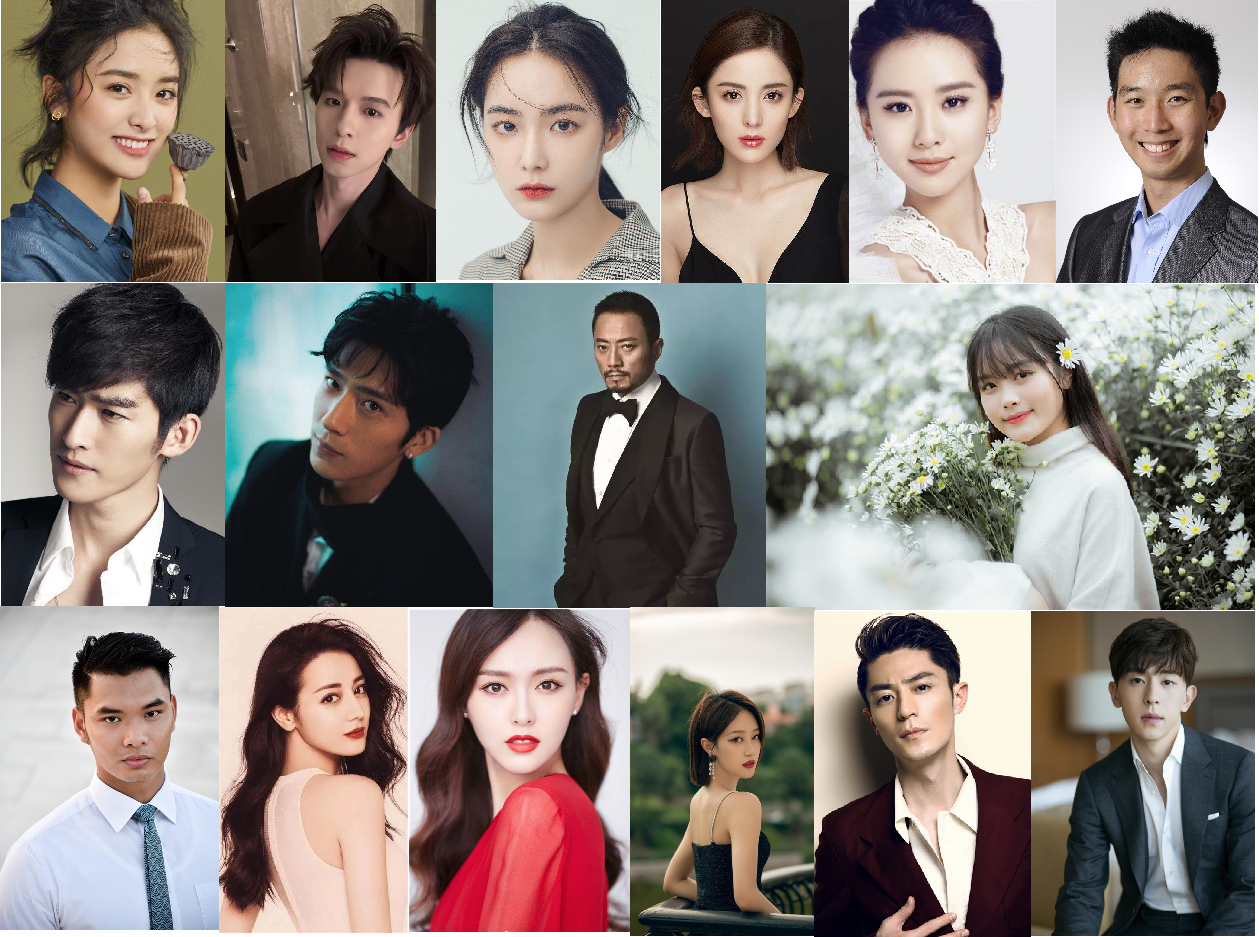}
    \caption{Examples of partial character ID images from ChineseID}
    \label{fig:chineseid}
\end{figure}

\begin{figure}[t]
    \centering
    \includegraphics[width=\linewidth]{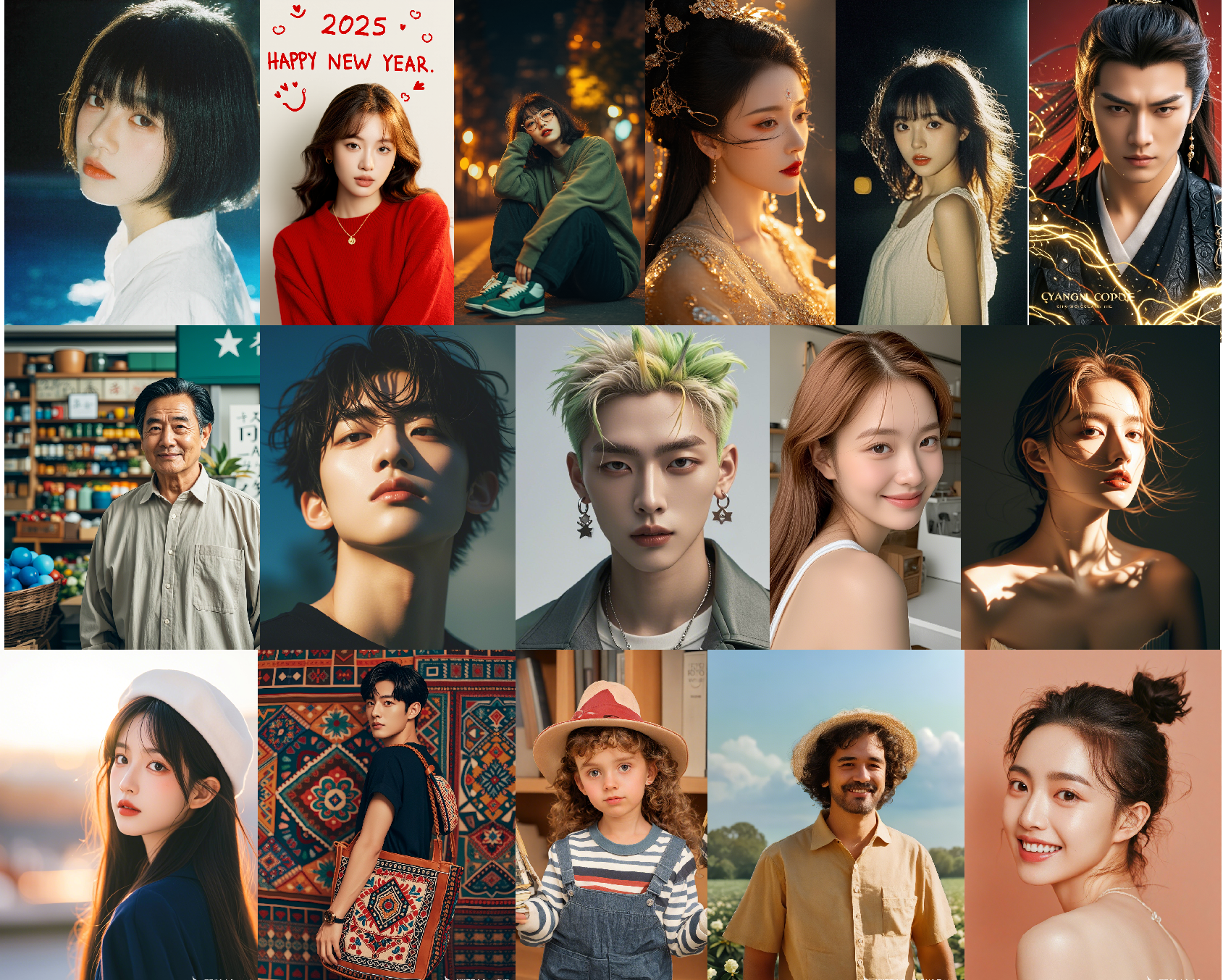}
    \caption{Examples of partial character ID images from GenerateID}
    \label{fig:generateid}
\end{figure}

We present partial data from Unsplash, ChineseID, and GenerateID in\cref{fig:unsplash} \cref{fig:chineseid} \cref{fig:generateid}.

\subsubsection{Prompts}
We display the prompt words for short prompt, editable long prompt, and manually collected prompt in the \cref{tab:short_prompt} \cref{tab:editable_long_prompt} \cref{tab:manually_collected_prompt} below.
\begin{table*}[t]
    \centering
    \makeatletter
    \def\@makecaption#1#2{%
        \vskip\abovecaptionskip
        \centering 
        \small #1: #2\par
        \vskip\belowcaptionskip
    }
    \makeatother
    \caption{Evaluation metric results from IBench on GenerateID with manually collected prompts}
    \label{tab:generateid_evaluation}
    \begin{tabular}{lccccccccc}
        \toprule
        Model & FID & Aesthetic & Image Quality & \multicolumn{3}{c}{Posediv} & Landmarkdiff & Exprdiv \\
        \cmidrule(lr){5-7}
        & & & & Yaw & Pitch & Roll & & \\
        \midrule
        Instantid & 61.13 & 0.568 & 0.422 & 24.23 & 13.94 & 12.60 & 0.125 & 0.541 \\
        PuLID(sdxl) & 31.24 & 0.659 & 0.490 & 22.41 & 11.58 & 12.74 & 0.107 & 0.669 \\
        PuLID(flux) & 43.64 & 0.697 & 0.461 & 20.29 & 12.14 & 12.19 & 0.099 & 0.574 \\
        EditID & 16.84 & 0.696 & 0.486 & 21.07 & 12.88 & 13.16 & 0.104 & 0.612 \\
        \midrule
        & Facesim & ClipI & ClipT & Dino & Fgis & & & \\
        \cmidrule(lr){2-6}
        Instantid & 0.184 & 0.699 & 0.219 & 0.071 & 0.099 \\
        PuLID(sdxl) & 0.372 & 0.832 & 0.251 & 0.113 & 0.206 \\
        PuLID(flux) & 0.393 & 0.803 & 0.238 & 0.128 & 0.211 \\
        EditID & 0.380 & 0.813 & 0.240 & 0.091 & 0.131 \\
        \bottomrule
    \end{tabular}
\end{table*}

\begin{table*}[t]
    \centering
    \makeatletter
    \def\@makecaption#1#2{%
        \vskip\abovecaptionskip
        \centering 
        \small #1: #2\par
        \vskip\belowcaptionskip
    }
    \makeatother
    \caption{Examples of partial prompts from short prompt}
    \begin{tabularx}{\textwidth}{cX} 
        \toprule
        \textbf{No.} & \textbf{Prompt} \\
        \midrule
        1 & Clothing portrait, a person wearing a spacesuit \\
        2 & Portrait, a person wearing a surgical mask \\
        3 & Background portrait, with a beautiful purple sunset at the beach in the background \\
        4 & Portrait, pencil drawing \\
        5 & Portrait, latte art in a cup \\
        6 & Portrait, side view, in papercraft style \\
        7 & Portrait, Madhubani, wearing a mask \\
        8 & Portrait, anime artwork \\
        9 & Portrait, energetic brushwork, bold colors, abstract forms, expressive, emotional \\
        10 & Portrait, a person wearing a doctoral cap \\
        \bottomrule
    \end{tabularx}
    \label{tab:short_prompt}
\end{table*}

\begin{table*}[t]
    \centering
    \makeatletter
    \def\@makecaption#1#2{%
        \vskip\abovecaptionskip
        \centering 
        \small #1: #2\par
        \vskip\belowcaptionskip
    }
    \makeatother
    \caption{Examples of partial prompts from editable long prompt}
    \begin{tabularx}{\textwidth}{c>{\raggedright\arraybackslash}X} 
        \toprule
        \textbf{No.} & \textbf{Prompt} \\
        \midrule
        1 & A solitary man, clad in a long, dark trench coat and a wide-brimmed hat, walks through a dimly lit alleyway, the only illumination coming from flickering street lamps casting elongated shadows. His footsteps echo softly against the cobblestones, creating a rhythmic pattern in the stillness of the night. The air is thick with mist, swirling around his silhouette, adding an air of mystery to his journey. Occasionally, he pauses, glancing over his shoulder, as if sensing an unseen presence. The distant sound of a train whistle punctuates the silence, enhancing the eerie, atmospheric setting of his solitary walk. \\
        2 & A young boy with tousled hair and a curious expression kneels in a sunlit garden, surrounded by vibrant blooms. He carefully places a delicate glass dome over a single, exquisite red rose, its petals glistening with morning dew. The sunlight filters through the glass, casting a kaleidoscope of colors onto the grass. His small hands gently adjust the dome, ensuring the rose is perfectly encased. The scene captures a moment of wonder and protection, as the boy admires the rose's beauty, the garden's lush greenery and colorful flowers providing a serene, enchanting backdrop. \\
        3 & A joyful child, wearing a bright yellow raincoat and red rubber boots, splashes gleefully in a series of puddles on a rainy day. The scene captures the child's infectious laughter as they jump, sending droplets flying in all directions. The overcast sky and gentle rain create a soothing backdrop, while the child's playful antics bring warmth and energy to the scene. As they stomp through the water, their reflection shimmers in the puddles, adding a magical touch. The child's carefree spirit and the rhythmic sound of raindrops create a heartwarming and lively atmosphere. \\
        4 & A young athlete, clad in a sleek black swimsuit and swim cap, stands at the edge of an Olympic-sized pool, the water shimmering under bright overhead lights. With a focused gaze, she adjusts her goggles, preparing for her training session. She dives gracefully into the water, her form streamlined and powerful, creating minimal splash. As she glides through the water, her strokes are precise and rhythmic, showcasing her dedication and skill. The camera captures her underwater, bubbles trailing behind her as she propels forward with determination. Finally, she emerges at the pool's edge, breathing deeply, her expression a mix of exhaustion and triumph. \\
        5 & A bearded man with a thoughtful expression stands in a cozy, dimly lit room filled with vintage decor, wearing a plaid shirt and jeans. He carefully selects a vinyl record from a wooden shelf lined with albums, the warm glow of a nearby lamp casting soft shadows. As he gently places the record onto the turntable, his fingers move with precision and care, reflecting his appreciation for music. The room is filled with the soft crackle of the needle touching the vinyl, and he closes his eyes momentarily, savoring the nostalgic sound. The ambiance is intimate, with the gentle hum of the record player and the soft lighting creating a serene atmosphere. \\
        6 & A passionate teacher stands at the front of a bright, modern classroom, holding a vibrant red marker in her hand, gesturing animatedly as she explains a complex concept to her attentive students. Her expression is one of enthusiasm and engagement, with her eyes sparkling with the joy of teaching. The whiteboard behind her is filled with colorful diagrams and notes, illustrating the topic at hand. Sunlight streams through large windows, casting a warm glow over the room, while students, seated at sleek desks, listen intently, some taking notes, others nodding in understanding, creating an atmosphere of dynamic learning and interaction. \\
        7 & A fierce woman stands confidently in her intricately detailed cosplay costume, embodying a warrior from a fantasy realm. Her armor, crafted from shimmering silver and deep blue materials, glistens under the ambient light, highlighting the ornate designs etched into the metal. Her long, flowing cape billows behind her as she strikes a powerful pose, her eyes focused and determined. The costume includes a helmet adorned with intricate patterns and a pair of gauntlets that suggest strength and agility. She holds a beautifully crafted sword, its blade reflecting the light, ready for battle. The background is a mystical landscape, with towering mountains and a sky painted in hues of twilight, enhancing the epic atmosphere of her warrior persona. \\
        8 & A seasoned actor, dressed in a vintage brown leather jacket, white shirt, and dark trousers, stands in a dimly lit, smoke-filled room, embodying a character from a noir film. His intense gaze and subtle smirk suggest a complex persona, while the shadows cast by a single overhead light add depth to his expression. As he moves, the camera captures his every nuanced gesture, from the flick of his wrist to the slight tilt of his head, conveying a sense of mystery and intrigue. The scene is set against a backdrop of old wooden furniture and a vintage rotary phone, enhancing the period atmosphere. \\
        \bottomrule
    \end{tabularx}
    \label{tab:editable_long_prompt}
\end{table*}

\begin{table*}[t]
    \centering
    \makeatletter
    \def\@makecaption#1#2{%
        \vskip\abovecaptionskip
        \centering 
        \small #1: #2\par
        \vskip\belowcaptionskip
    }
    \makeatother
    \caption{Examples of partial prompts from manually collected prompt}
    \begin{tabularx}{\textwidth}{c>{\raggedright\arraybackslash}X} 
        \toprule
        \textbf{No.} & \textbf{Prompt} \\
        \midrule
        1 & Person B, surprised, flying car, backyard, daytime, casual outfit, excitement, medium shot, A young Black woman, dressed in simple, comfortable home attire, steps out of her house into the bright daylight. She gazes in awe at a futuristic flying car gracefully descending in her backyard, close to the landing point. Her expression transforms from astonishment to sheer excitement, capturing the moment perfectly. The scene is set in a modern American neighborhood, with elements of advanced technology like sleek flying vehicles and high-tech devices enhancing the atmosphere. The composition is a medium shot, focusing on her joyful reaction amidst the backdrop of her home. \\
        2 & Young female singer, microphone, concert stage, retro neon lights, posters, evening, shiny silver jacket, skinny jeans, high heels, curly hair, dynamic atmosphere, mid-shot, eye level, A vibrant young female singer stands confidently in front of a microphone, prepared to captivate the audience with her performance. She is on a lively concert stage adorned with dazzling retro neon lights and colorful posters, evoking a nostalgic ambiance. It is a bustling evening, and she is dressed in a striking shiny silver jacket that glimmers under the lights, complemented by tight-fitting skinny jeans and stylish high heels. Her hair is voluminous and curly, adding to her energetic presence. The stage lights pulse and flicker in sync with the upbeat music, enhancing the dynamic atmosphere of the scene. The composition is framed as a mid-shot from an eye-level perspective, inviting viewers into the exhilarating moment of the performance. \\
        3 & Young woman, self-service check-in system, modern office, morning, professional attire, tablet, busy staff, large screen, real-time data, A focused young Chinese woman with long hair, dressed in professional attire, is operating a sleek tablet at a self-service check-in kiosk. The scene is set in a contemporary office filled with busy staff moving around her. It's 9 AM in the morning, with natural light streaming through large windows. In the background, a large screen displays real-time data, adding a high-tech feel to the atmosphere. The composition is a medium shot at eye level, capturing the dynamic environment and the woman's concentration on her task. \\
        4 & Animated character, cashier interaction, shopping process, afternoon, friendly cashier, crowd of customers, A cheerful animated character with large, expressive eyes and a stylish outfit, handing selected items to a friendly cashier at a brightly lit checkout counter. The character is smiling while making a payment, and the cashier is engaging in friendly conversation. In the background, a diverse crowd of customers, including Asian individuals, patiently waits in line. The scene takes place in the afternoon, with warm sunlight streaming through the store's windows, creating a lively and inviting atmosphere. \\
        5 & Li Bai, courtyard, well railing, poetry, candlelight, night, warm light, glowing words, close-up, eye level, A young Chinese poet, Li Bai, dressed in a soft moon-white robe, stands in a serene traditional courtyard at night, his face warmly illuminated by the gentle flicker of candlelight. He gazes thoughtfully at the well railing, deeply immersed in reciting poetic verses. Each word he utters releases a faint glow that softly drifts into the air. The ancient courtyard is adorned with lush green pine trees and a bright, shining moon above, adding to the tranquil and poetic atmosphere. The scene is captured in a close-up, eye-level perspective, evoking a sense of intimacy and reflection. \\
        6 & Young man, park bench, Shanghai, evening, dark coat, jeans, lost expression, skyline, wilted trees, A young Chinese man wearing a dark coat and jeans, sitting alone on a weathered park bench in a quiet park in Shanghai during the evening. He has slightly disheveled hair and a lost expression in his eyes, reflecting deep thoughts. The background features a few wilted trees and a distant city skyline, creating a serene yet melancholic atmosphere. The scene is captured in a mid-range view, at eye level, emphasizing his solitude and contemplation as the soft evening light casts gentle shadows around him. \\
        7 & Student, desk, books, tablet, modern study room, evening, holographic projection, relaxed smile, A Chinese student with black hair and glasses, dressed in a simple t-shirt and jeans, sits at a modern wooden desk cluttered with books and notes, appearing tired yet focused. The background features a contemporary study room illuminated by a desk lamp, with a study plan hanging on the wall. It's evening, casting a warm glow. The tablet displays a holographic projection of the answers, and the student’s face lights up with a relaxed smile, reflecting a sense of relief and understanding. The scene is captured in a mid-shot from a slightly high angle, creating a thoughtful and engaging atmosphere. \\
        8 & Female protagonist, crowd, smiling, waving, golden qipao, high bun, red hairpin, elegant movement, stage background, close-up, eye-level, A Chinese female protagonist gracefully walks out of a vibrant crowd, smiling warmly and waving. She is dressed in a stunning golden traditional qipao, her hair elegantly styled in a high bun, accentuated with a striking red hairpin. The backdrop features a dazzling stage, illuminated by focused lights that highlight her elegant movement. The composition is a close-up shot at eye-level, capturing the essence of her joyful demeanor amidst the festive atmosphere. \\
        \bottomrule
    \end{tabularx}
    \label{tab:manually_collected_prompt}
\end{table*}

\subsection{More Results}
\label{subsec:More Results}
\subsubsection{Unsplash Experimental Results}
This paper focuses more on long prompts, but IBench itself also includes an evaluation for short prompts. When compared with the ChineseID editable long prompt and GenerateID Typemovie prompt, our evaluation metrics on the Unsplash short prompt typically show lower facesim (face similarity) and higher posediv (pose diversity), which aligns with general variation patterns. However, as shown in \cref{fig:unsplash} and \cref{tab:short_prompt}, the large variations in portrait angles in Unsplash, combined with overly short prompts, lead to generally poor model performance. Due to the coarse text control of short prompts, we incorporated a [man/woman] design in the prompts. When the model correctly identifies the gender of the image, it replaces "person" in the prompt accordingly.

\subsection{Supplementary Experimental Materials}
\label{subsec:More Results}
Prompts from \cref{fig:shift_strategies} and \cref{fig:fusion_methods}:
\begin{itemize}
    \item A figure in a contemplative stance, partially turned away from the viewer, dressed in a composed outfit that enhances his serene demeanor. The gentle play of light highlights the subtle expressions on his face, capturing the delicate contours of his jawline and the graceful curve of his neck in a side view. His profile is framed by the tranquil setting in the background, which complements the mood of introspection.
    \item This is a young Asian individual with fair skin, whose straight, deep brown hair falls just below the shoulders. Thick bangs partially cover the forehead, accentuating a sharply defined side profile and large, gentle eyes that exude a hint of tenderness. She is slightly turning, with her shoulder gently tilted, displaying an elegantly natural pose as if she is slowly turning to admire the lush green scenery around her. Her high, refined nose stands out in the soft sunlight, and her slightly parted lips reveal a serene yet enchanting smile. Sunlight filters through the dense branches and leaves, casting dappled shadows on her profile and infusing the entire scene with a gentle, peaceful ambiance.
\end{itemize}

\subsection{Show case of Some EditID Results}
\begin{figure*}[t]
    \centering
    \includegraphics[width=\textwidth]{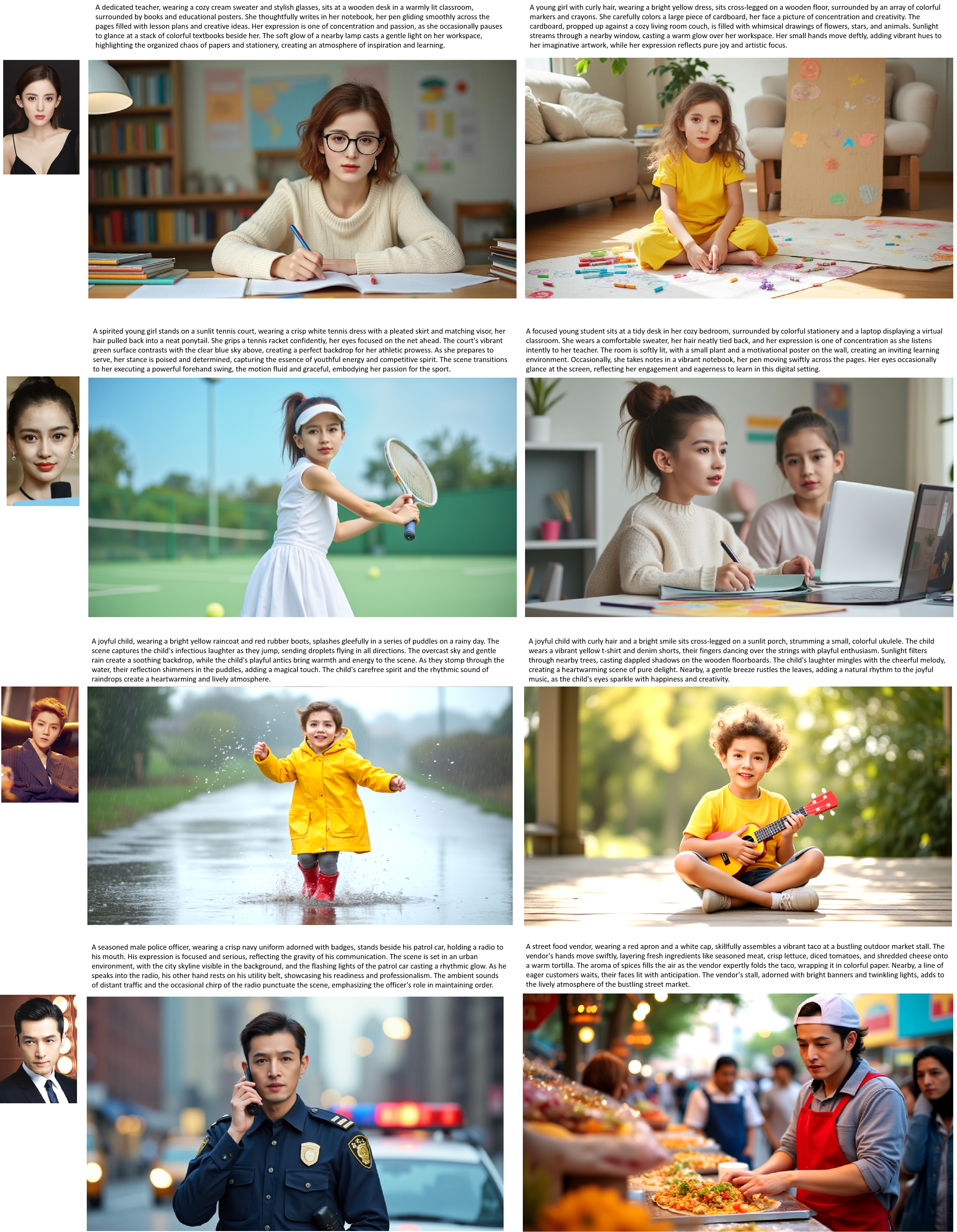}
    \caption{Some results of EditID}
    \label{fig:figure15}
\end{figure*}

\cref{fig:figure15} shows some results of editid.As shown in the image, for the input ID, not only can it display facial and body movements, but it can also alter attributes such as age and hairstyle through prompts, demonstrating exceptional editability. Moreover, the quality of the generated images is extremely high.
